\pdfoutput=1

\documentclass[11pt]{article}

\usepackage[]{EMNLP2023}

\usepackage{times}
\usepackage{latexsym}

\usepackage[T1]{fontenc}

\usepackage[utf8]{inputenc}

\usepackage{microtype}

\usepackage{inconsolata}

\usepackage{CJK}                
\usepackage{multirow}           
\usepackage{booktabs}           
\usepackage{arydshln}
\usepackage{graphicx}           
\usepackage{float}              
\usepackage{bm}                 
\usepackage[tikz]{bclogo}       
\usepackage{amsmath}            
\usepackage{amsfonts}           
\usepackage{pgfplots}           
\pgfplotsset{compat=1.18}
\usepackage{caption}
\usepackage{subcaption}         

\usepackage[hang,flushmargin]{footmisc} 

\definecolor{brickred}{HTML}{b92622}
\definecolor{midnightblue}{HTML}{005c7f}
\definecolor{salmon}{HTML}{f1958d}
\definecolor{burntorange}{HTML}{f19249}
\definecolor{junglegreen}{HTML}{4dae9d}
\definecolor{forestgreen}{HTML}{499c5e}
\definecolor{pinegreen}{HTML}{3d8a75}
\definecolor{seagreen}{HTML}{6bc1a2}
\definecolor{limegreen}{HTML}{97c65a}

\newcommand{\finding}[1]{
\begin{bclogo}[couleur= black!05, epBord= 1, arrondi=0.1, logo=\bclampe, marge= 2, ombre=true, blur, couleurBord=black!10, tailleOndu=3, sousTitre ={\em #1}]{} 
\end{bclogo}
}

\title{
CLEME: Debiasing Multi-reference Evaluation for \\ Grammatical Error Correction
}

\author{
Jingheng Ye$^{1}$\thanks{ $^*$ indicates equal contribution.},~Yinghui Li$^{1*}$,~Qingyu Zhou$^{2}$,~Yangning Li$^{1,3}$,\\~\textbf{Shirong Ma}$^{1}$,~\textbf{Hai-Tao Zheng}$^{1,3}$\thanks{ $^{\dagger}$ Corresponding author: Hai-Tao Zheng. (E-mail: zheng.haitao@sz.tsinghua.edu.cn)}, \textbf{Ying Shen}$^{4}$\\
        $^{1}$Tsinghua Shenzhen International Graduate School, Tsinghua University \\ 
        $^{2}$OPPO Research Institute, $^{3}$Peng Cheng Laboratory \\
        $^{4}$School of Intelligent Systems Engineering, Sun-Yat Sen University\\
        \texttt{\{yejh22,liyinghu20\}@mails.tsinghua.edu.cn}
}

\begin{document}

\maketitle

\begin{abstract}
Evaluating the performance of Grammatical Error Correction (GEC) systems is a challenging task due to its subjectivity. 
Designing an evaluation metric that is as objective as possible is crucial to the development of GEC task.
However, mainstream evaluation metrics, i.e., reference-based metrics, introduce bias into the multi-reference evaluation by extracting edits without considering the presence of multiple references.
To overcome this issue, we propose \textbf{C}hunk-\textbf{LE}vel \textbf{M}ulti-reference \textbf{E}valuation (\textbf{CLEME}), designed to evaluate GEC systems in the multi-reference evaluation setting.
CLEME builds chunk sequences with consistent boundaries for the source, the hypothesis and references, thus eliminating the bias caused by inconsistent edit boundaries.
Furthermore, we observe the consistent boundary could also act as the boundary of grammatical errors, based on which the F$_{0.5}$ score is then computed following the correction independence assumption.
We conduct experiments on six English reference sets based on the CoNLL-2014 shared task.
Extensive experiments and detailed analyses demonstrate the correctness of our discovery and the effectiveness of CLEME.
Further analysis reveals that CLEME is robust to evaluate GEC systems across reference sets with varying numbers of references and annotation styles~\footnote{
All the source codes of CLEME are released at~\url{https://github.com/THUKElab/CLEME}.
}.

\end{abstract}

\section{Introduction}\label{sec:intro}

Grammatical Error Correction (GEC) is a task that involves making \textit{local substitutions} to correct grammatical errors in a given ungrammatical text~\cite{bryant2022grammatical, ma-etal-2022-linguistic, DBLP:journals/corr/abs-2210-12692, DBLP:journals/corr/abs-2306-17447}. 
The practical value of GEC in daily life has led to increasing attention being paid to this task~\cite{li-etal-2021-miss,li-etal-2022-learning-dictionary,li-etal-2022-past,kaneko-etal-2022-interpretability, DBLP:journals/corr/abs-2307-09007, 10095675}.
However, it is intractable to evaluate GEC systems due to the highly subjective nature of the task and the low inter-annotator agreement (IAA)~\cite{choshen-abend-2018-automatic}.
Therefore, most datasets improve compatibility by incorporating multiple references to guarantee a more realistic evaluation of the model performance.

\begin{figure}[tbp!]
\centering
\includegraphics[width=\linewidth]{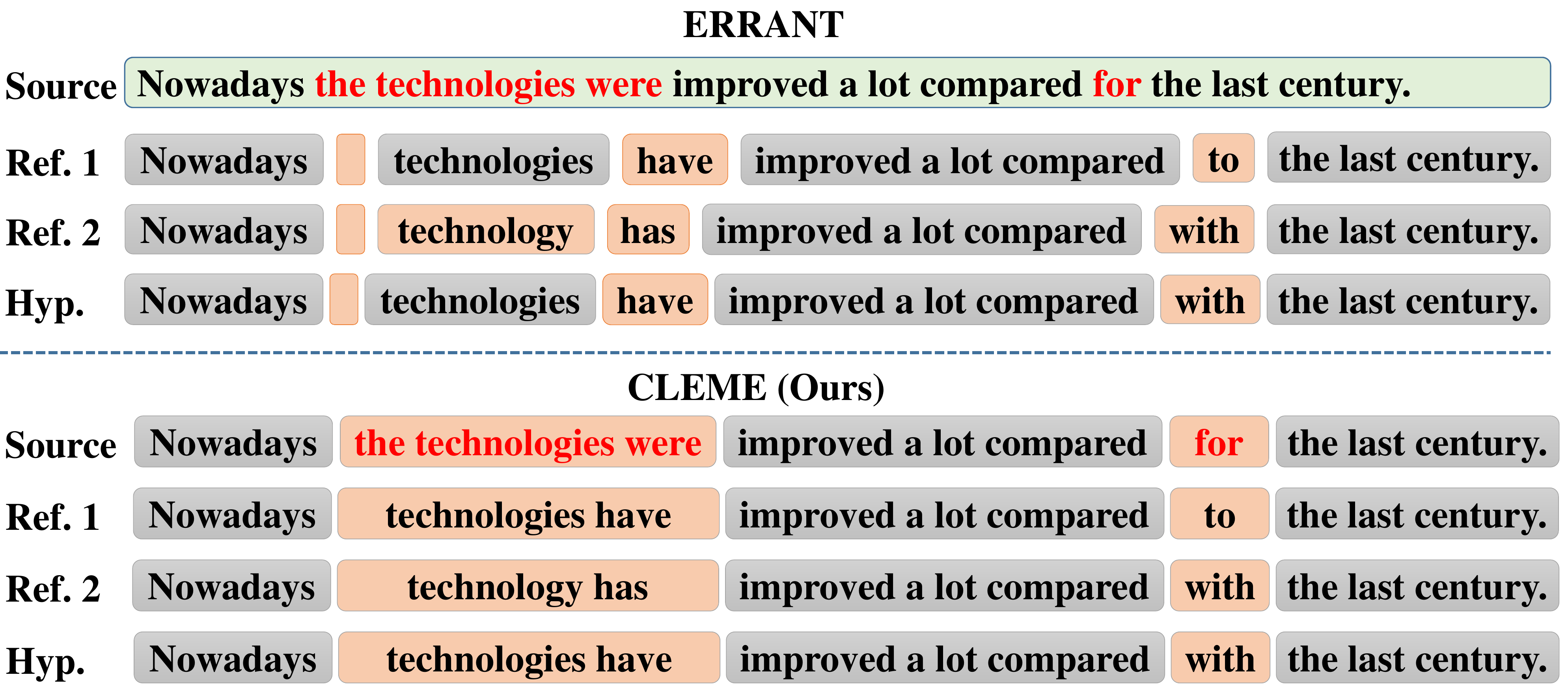}
\caption{A comparison of edits automatically extracted by ERRANT and \textbf{CLEME}. An \textcolor{orange}{orange} block is an edit.}
\label{fig:intro}
\end{figure}

There are two broad categories of GEC metrics: reference-based and reference-less.
Reference-based metrics evaluate GEC systems by comparing their hypotheses and human-annotated references in terms of edits~\cite{dahlmeier-ng-2012-better,bryant-etal-2017-automatic} or n-grams~\cite{napoles-etal-2015-ground}.
Reference-less metrics are proposed to evaluate GEC systems without references.
However,~\citet{deutsch2022limitations} demonstrate that reference-less metrics are inherently biased and limited in their ability to evaluate generated text.
Therefore, we focus on reference-based metrics, which can evaluate in an interpretable manner, thus providing useful insights for model analysis.

Figure~\ref{fig:intro} illustrates how existing reference-based metrics, such as ERRANT, extract the edit and then compute the F$_{0.5}$ score by comparing hypotheses and references.
However, these metrics often fail to consider multiple references, which can result in bias during multi-reference evaluation.
We argue that this bias arises because the current approach rewards equally good corrections unfairly.
For instance, the ungrammatical phrase \textit{the technologies were} is equally well-corrected by both Ref. 1 and Ref. 2.
However, if a hypothesis aligns with Ref. 1's corrections (i.e., [\textit{the} $\to \epsilon$] and [\textit{were} $\to$ \textit{have}], TP=2), it will be rewarded less than the corrections of Ref. 2 (i.e., [\textit{the} $\to\epsilon$], [\textit{technologies} $\to$ \textit{technology}] and [\textit{were} $\to$ \textit{has}], TP=3).

In this paper, we propose \textbf{C}hunk-\textbf{LE}vel \textbf{M}ulti-reference \textbf{E}valuation (\textbf{CLEME}), which enables unbiased F$_{0.5}$ scores for GEC multi-reference evaluation.
Inspired by~\cite{gotou-etal-2020-taking}, CLEME transforms the source, the hypothesis and all the references into chunk sequences with consistent boundaries, thereby eliminating the bias in GEC multi-reference evaluation.

Existing metrics assume that corrections of grammatical errors are dependent.
That is, whenever there is more than one reference for a source, the metrics try each reference in turn, and then the highest score is taken as the final score.
However, we observe that grammatical errors corrections in terms of chunks can be considered \textbf{approximately independent}.
For example, the ungrammatical phrases \textit{the technologies were} and \textit{for} shown in Figure~\ref{fig:intro} can be corrected independently, i.e., the correction of \textit{the technologies were} has no bearing on the correction of \textit{for}.
Based on this observation, we compute F$_{0.5}$ scores following the assumption that corrections of grammatical errors are independent.
Specifically, we iterate through the chunks of a hypothesis and consider a chunk correct if it matches any of the corresponding chunks in the references.
In this case, the hypothesis in Figure~\ref{fig:intro} would be rewarded 2TP, rather than 1TP and 1FP, which is the traditional case.
To demonstrate the effectiveness and robustness of CLEME, we conduct experiments on six English reference sets with varying numbers of references and annotation styles, either calculating the F$_{0.5}$ score at the corpus- or sentence-level.

In summary, our contributions are three folds:
\begin{itemize}
    \item [(1)] We propose CLEME, a reference-based metric that evaluates GEC systems at the chunk-level, aiming to provide unbiased F$_{0.5}$ scores for GEC multi-reference evaluation.
     
    \item [(2)] We observe that the corrections of grammatical errors in terms of chunks are approximately independent.
    Therefore, we propose to compute F$_{0.5}$ scores based on the correction independence assumption.

    \item [(3)] Extensive experiments and human evaluation are conducted to confirm the effectiveness and robustness of our approach.
\end{itemize}

\section{Preliminary Study}


\subsection{Consistent Boundaries}\label{subsec:chunk-partition}
We determine consistent chunk-level boundaries by chunk partition process to debias the multi-reference evaluation, as depicted in Figure~\ref{fig:cleme}.
We first extract the edit sets of the hypothesis and references, and then merge the overlapping edits into a chunk.
It's worth noting that the source, hypothesis and references are all segmented into \textit{chunk sequences} with the same number of chunks, regardless of the number of their tokens.
This process is straightforward since we can locate and examine all possible corrections of an erroneous chunk.
For example, the chunk \textit{by the} can be corrected in two ways, i.e., \textit{with} in Ref. 1 and \textit{through} in Ref. 2.
The resulting chunks fall into three categories:
1) \textbf{unchanged chunks}, which contain the same text segments as the source sentence,
2) \textbf{corrected chunks}, which consist of non-empty text segments different from the source sentence, and
3) \textbf{dummy chunks} are empty chunks.


\subsection{Boundaries of Grammatical Errors}
\label{subsec:chunk-boundary}
\begin{figure*}[ht!]
\centering
\includegraphics[scale=0.302]{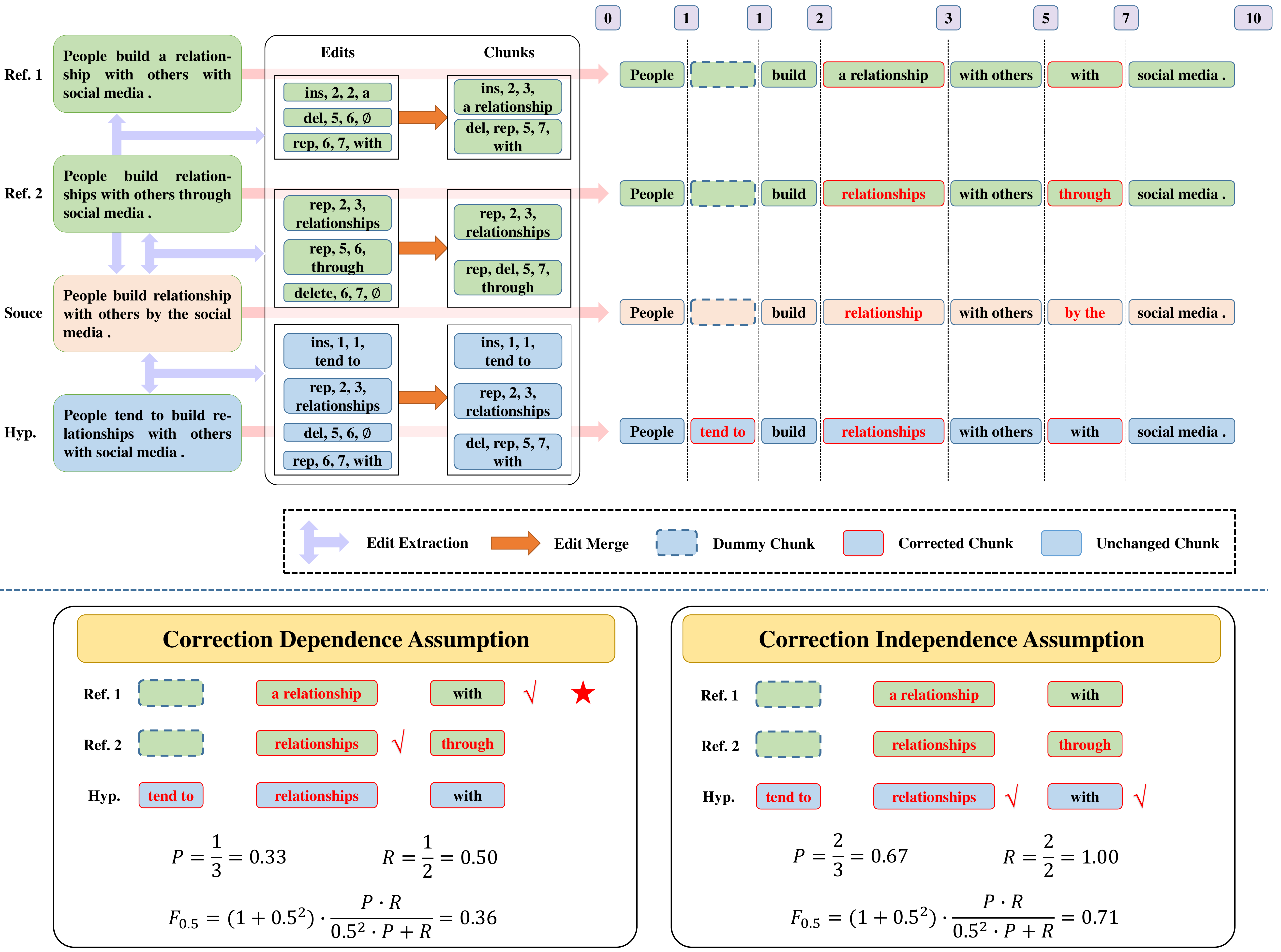}
\caption{
Overview of our approach CLEME. 
CLEME first
1) extracts edits of the hypothesis and the references, 
2) merges the overlapping edits into chunks, and then 
3) computes the F$_{0.5}$ scores based on two different assumptions.
}
\label{fig:cleme}
\end{figure*}

Figure~\ref{fig:cleme} illustrates the merging of overlapping edits into either corrected or dummy chunks, which are then separated by unchanged chunks.
This raises the question, \emph{are chunk boundaries the boundaries of grammatical errors?}


\paragraph{\textbf{Dataset}.}
To answer the question, we conduct experiments on BN-10GEC~\cite{bryant-ng-2015-far}. 
The dataset comprises 1,312 source sentences that are identical to the CoNLL-2014 test data~\cite{ng-etal-2014-conll}.
Each source sentence is associated with 10 references annotated by 10 native English speakers, including two official annotators of CoNLL-2014, the first author of the paper, and seven freelancers recruited via an online recruitment website.

\paragraph{\textbf{Experiment Setup}.}
For each source sentence, we sample 9 references and run the chunk partition process described in Section~\ref{subsec:chunk-partition}.
The resulting chunk sequences are determined collectively by all 9 references.
The edits of the remaining reference $\{e_1,\cdots,e_M\}$ are then used to calculate the following three statistics:
1) The In-Corrected-Chunk (ICC) ratio indicates the proportion of edits included by corrected/dummy chunks of the other references.
An edit is included by a chunk if the interval of the edit falls within that of the chunk.
2) The In-Unchanged-Chunk (IUC) ratio gives the proportion of edits included by unchanged chunks of the other references.
3) The Cross-Chunk (CC) ratio computes the proportion of edits that extend beyond the original boundaries.
These statistics are calculated as follows:

\begin{equation}\small\begin{aligned}
\text{ICC}=\frac{1}{M}\sum^M_{i=1}f_1(e_i),
\end{aligned}\end{equation}

\begin{equation}\small\begin{aligned}
\text{IUC}=\frac{1}{M}\sum^M_{i=1}f_2(e_i),
\end{aligned}\end{equation}

\begin{equation}\small\begin{aligned}
\text{CC}=1-\text{ICC}-\text{IUC},
\end{aligned}\end{equation}
where $M$ is the number of edits from the remaining reference.
If the edit $e_i$ is included in a corrected/dummy chunk, the function $f_1(e_i)$ returns 1, otherwise 0.
Likewise, if the edit $e_i$ is included in an unchanged chunk, the function $f_2(e_i)$ returns 1, otherwise 0.
We sample 9 different references for chunk partition in each run and repeatedly calculate the statistics using the remaining reference.

\paragraph{\textbf{Results}.}
As shown in Table~\ref{tab:boundary}, the number of corrected and dummy chunks are less than that of edits since overlapping edits are merged into a chunk.
A total of 90.66\% edits are included by the corrected/dummy chunks, which suggests the grammatical errors to be corrected have been considered by the other references.
However, only 7.74\% edits are included by corrected chunks, indicating that these edits may be over-corrected since the other references believe no grammatical errors needed correction. 
Interestingly, 1.61\% edits cross the chunk boundaries, suggesting that the chunk boundaries are stable enough to serve as the boundaries of grammatical errors to some extent. Additionally, human evaluation in Section~\ref{subsec:human_evaluation} could be used as another argument to support this conclusion.
Therefore, we have the following assumption.

\begin{table}[tbp!]
\centering
\scalebox{0.8}{
\begin{tabular}{llc}
\toprule
\bf{Item} & \bf {Number (perc.)} & \bf{Length} \\
\midrule

Sentences  & 1,312  & 23.0 \\
References & 13,120 & 22.9 \\
Edits      & 36,677 & 1.0  \\
Unchanged Chunks       & 93,469 (77.63\%) & 2.5 \\
Corrected/Dummy Chunks & 26,948 (22.37\%) & 2.4 \\
ICC & 33,251 (90.66\%) & -\\
IUC & 2,837 (7.74\%)   & -\\
CC  & 589 (1.61\%)     & -\\

\bottomrule
\end{tabular}}
\caption{
Statistics of the BN-10GEC dataset.
}\label{tab:boundary}
\end{table}

\finding{
Correction independence assumption:\newline
\scalebox{0.92}{grammatical error corrections are independent.}
}\label{finding:assumption}

That is, the correction of a grammatical error does not impact the correction of other grammatical errors. With this assumption, F$_{0.5}$ scores can be calculated using an alternate method, which will be introduced in Section~\ref{sec:method}.

\section{Method}\label{sec:method}
\subsection{Chunk Evaluation}
As shown in Figure~\ref{fig:cleme}, each chunk consists of edit operation(s), start index, end index, and correct tokens.
Conventional reference-based metrics such as MaxMatch (M$^2$) and ERRANT compute F$_{0.5}$ scores based on the correction dependence assumption.
They evaluate the performance for each reference separately and select the one that yields the best result for the source sentence.
\textbf{CLEME-dependent} also computes F$_{0.5}$ scores in this way by treating corrected/dummy chunks as edits.
On the other hand, \textbf{CLEME-independent} is proposed to compute F$_{0.5}$ scores based on the correction independence assumption.
A corrected/dummy chunk from the hypothesis is considered correct if it matches one of the corresponding chunks from the references.
It is worth noting that CLEME is able to fully inherit pre-classified errors from ERRANT, where each corrected/dummy chunk may consist of multiple error with different types.

\subsection{Length Weighting}\label{subsec:length-weighting}
The average length of chunks is much longer than that of edits shown in Table~\ref{tab:boundary}, resulting in the unfairness of chunk evaluation if a longer chunk is rewarded equally with a shorter one.
Therefore, we add length weighting to the chunk evaluation.
The intuition of length weighting is to compensate for long chunk matching.
The weights of True Positives (TPs), False Positives (FPs), and False Negatives (FNs) are computed as follows:\footnote{We do not apply length weighting to TNs since it is unnecessary for F$_{0.5}$ scores.}

\begin{figure}[tb!]
\centering
\scalebox{0.70}{
\begin{tikzpicture}
\begin{axis}[
    legend pos=outer north east,
    xlabel=Length of Chunk $x$,
    ylabel=Length Weight $w$,
    ylabel style = {yshift=0pt},
    xmin=0, xmax=6.0,
    ymin=0, ymax=3.0,
    xtick={0,1,2,3,4,5,6},
    ytick={0.0,0.5,1.0,1.5,2.0,2.5,3.0},
    yticklabels={0.0,0.5,1.0,1.5,2.0,2.5,3.0},
]
\addplot[
    domain=0:6,
    color=blue!80,thick,
] {2/(1+1*exp(2-x))};
\addlegendentry{$\alpha_1=2$}

\addplot[
    domain=0:6,
    color=green,thick,
] {3/(1+2*exp(2-x))};
\addlegendentry{$\alpha_1=3$}

\addplot[
    domain=0:6,
    color=red!80,thick,
] {5/(1+4*exp(2-x))};
\addlegendentry{$\alpha_1=5$}

\addplot[
    domain=0:6,
    color=violet!80,thick
] {2/(1+1*exp(x-2))};
\addlegendentry{$\alpha_2=2$}

\addplot[
    domain=0:6,
    color=orange!80,thick
] {3/(1+2*exp(x-2))};
\addlegendentry{$\alpha_2=3$}

\addplot[
    domain=0:6,
    color=gray,thick
] {5/(1+4*exp(x-2))};
\addlegendentry{$\alpha_2=5$}

\addplot[
    domain=0:6,
    color=black,dashed,thick,
] {0.25};

\addplot[
    domain=0:6,
    color=black,dashed,thick,
] {2.50};

\end{axis}
\end{tikzpicture}}
\caption{Curves of length weighting with different $\alpha$ for $\ell=2$. All the curves pass through the point $(\ell,1.0)$. A curve with a larger scale factor has a greater slope.}
\label{fig:length-weighting}
\end{figure}
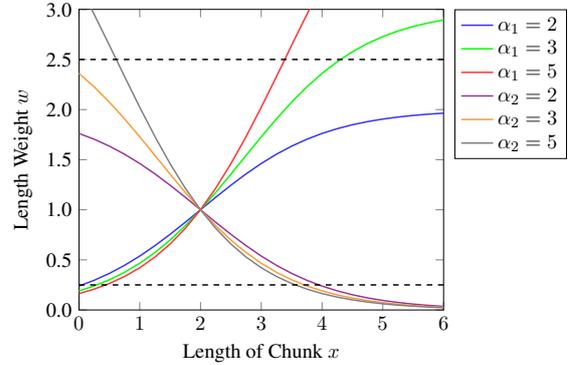

\begin{small}\begin{equation}\begin{aligned}
    w^{\text{TP}}=\text{clip}\left(\frac{\alpha_1}{1+(\alpha_1-1)\exp(\ell-x)}, c_{\text{min}}, c_{\text{max}}\right), 
\end{aligned}\end{equation}\end{small}
\begin{small}\begin{equation}\begin{aligned}
    w^{\text{FP}}=\text{clip}\left(\frac{\alpha_2}{1+(\alpha_2-1)\exp(x-\ell)}, c_{\text{min}}, c_{\text{max}}\right),
\end{aligned}\end{equation}\end{small}
\begin{small}\begin{equation}\begin{aligned}
    w^{\text{FN}}=\text{clip}\left(\frac{\alpha_3}{1+(\alpha_3-1)\exp(\ell-x)}, c_{\text{min}}, c_{\text{max}}\right),
\end{aligned}\end{equation}\end{small}
where $\alpha_1$, $\alpha_2$ and $\alpha_3$ are scale factors for TPs, FPs and FNs respectively, $x$ is the length of the chunk, $\ell$ is the average length of chunks, and the function $\text{clip}(v,a,b)$ clips the value $v$ between $a$ and $b$.
The curves of length weighting are depicted in Figure~\ref{fig:length-weighting}.
Formally, given a system corrected/dummy chunk set $\boldsymbol{C}^{\boldsymbol{H}}$ and a gold corrected/dummy chunk set $\boldsymbol{C}^{\boldsymbol{R}}$, we apply length weighting on each chunk to compute precision, recall and F$_{0.5}$ as follows:

\begin{normalsize}\begin{equation}\small
P=\frac{\sum\limits_{c\in\boldsymbol{C}^{\boldsymbol{H}}\cap\boldsymbol{C}^{\boldsymbol{R}}}w_c^{\text{TP}}}{\sum\limits_{c\in\boldsymbol{C}^{\boldsymbol{H}}\cap\boldsymbol{C}^{\boldsymbol{R}}}w_c^{\text{TP}} + \sum\limits_{c\in\boldsymbol{C}^{\boldsymbol{H}}\setminus\boldsymbol{C}^{\boldsymbol{R}}}w_c^{\text{FP}}},
\end{equation}\end{normalsize}
\begin{normalsize}\begin{equation}\small
R=\frac{\sum\limits_{c\in\boldsymbol{C}^{\boldsymbol{H}}\cap\boldsymbol{C}^{\boldsymbol{R}}}w_c^{\text{TP}}}{\sum\limits_{c\in\boldsymbol{C}^{\boldsymbol{H}}\cap\boldsymbol{C}^{\boldsymbol{R}}}w_c^{\text{TP}} + \sum\limits_{c\in\boldsymbol{C}^{\boldsymbol{R}}\setminus\boldsymbol{C}^{\boldsymbol{H}}}w_c^{\text{FN}}},
\end{equation}\end{normalsize}
\begin{equation}\small
    F_\beta = (1+\beta^2)\cdot\frac{P\cdot R}{(\beta^2\cdot P)+ R},
\label{eq:F}\end{equation}
where $\beta=0.5$ is usually used, which weighs precision twice as much as recall.

\subsection{Corpus-level v.s. Sentence-level}
We compute F$_{0.5}$ scores of GEC systems at both corpus-level and sentence-level following~\cite{gong2022revisiting}.
Corpus-level metrics compute an F$_{0.5}$ score over the entire dataset.
Sentence-level metrics compute an F$_{0.5}$ score over each sentence of the dataset and evaluate GEC systems by using the average F$_{0.5}$ score.
CLEME-dependent and CLEME-independent are corpus-level metrics, and their sentence-level variants are respectively \textbf{SentCLEME-dependent} and \textbf{SentCLEME-independent}.
Both levels of the GEC metric are developed to provide more user-friendly options.
Sentence-level metrics should be used if consistent evaluation weight for each sample is desired.
This ensures that the evaluation result of each sample has the same influence on the final score.
On the other hand, if harder samples containing more edits should have larger weight, then corpus-level metrics should be used instead.

\section{Experiments}
\subsection{Correlations with Human Judgments}

\paragraph{\textbf{Dataset}.}
To verify the effectiveness of CLEME, we measure correlations between reference-based metrics and human judgments on multiple English reference sets, including CoNLL-2014~\cite{grundkiewicz2015human}, BN-10GEC~\cite{bryant-ng-2015-far} and SN-8GEC~\cite{sakaguchi-etal-2016-reassessing}.
All the reference sets are based on CoNLL-2014~\cite{ng-etal-2014-conll}, consisting of 1,312 source sentences.
SN-8GEC collected 8 references sets of annotations from both experts and non-experts, including 4 sets of minimal edits and 4 sets of fluency edits (2 by experts and 2 by non-experts).
Reference sets statistics are reported in Appendix~\ref{subsec:statistics-ref}.

The human judgments for the outputs of 13 GEC systems (including the unchanged source text) are presented by ~\cite{grundkiewicz2015human}, where eight native speaker were asked to rank the output of all the systems from best to worst.
Two system ranking lists are generated using Expected Wins (EW)~\cite{machacek-bojar-2013-results} and TrueSkill (TS)~\cite{sakaguchi-etal-2014-efficient} respectively.

\paragraph{\textbf{Experiment Settings}.}
Following~\cite{gong2022revisiting,chollampatt-ng-2018-reassessment}, we compute the Pearson $\gamma$ and Spearman correlation coefficient $\rho$ between reference-based metrics and human judgments based on corpus-level ranking.
We tune the hyperparameters on CoNLL-2014 and keep the hyperparameters on the other reference sets, in order to demonstrate the adaptability of our approach.
The detailed hyperparameters of CLEME are reported in Appendix~\ref{subsec:hyperparameters}.

\paragraph{\textbf{Evaluation Metrics}.}
We compare our approach with the following reference-based metrics, including corpus- and sentence-level variants~\footnote{
We do not experiment with I-measure~\cite{felice-briscoe-2015-towards} due to its negative correlation and high computing complexity~\cite{grundkiewicz2015human}.}:

\begin{itemize}
\item \textbf{GLEU} and \textbf{SentGLEU}~\cite{napoles-etal-2015-ground} are n-gram based metrics, which reward hypothesis n-grams that overlap with the reference but not the source and penalize hypothesis n-grams that overlap with the source but not the reference.

\item \textbf{M$^2$} and \textbf{SentM$^2$}~\cite{dahlmeier-ng-2012-better} dynamically extract the hypothesis edits with the maximum overlap of gold annotations.

\item \textbf{ERRANAT} and \textbf{SentERRANT}~\cite{bryant-etal-2017-automatic} extract edits by utilizing a linguistically-enhanced alignment algorithm.

\item \textbf{PT-M$^2$} and \textbf{SentPT-M$^2$}~\cite{gong2022revisiting} are recently proposed reference and PLM-based GEC metric, which score edits using the knowledge of pre-trained language model.

\end{itemize}

Additionally, CLEME can evaluate GEC systems by accuracy scores, which is usually not implemented by conventional reference-based metrics.
Please refer to Appendix~\ref{subsec:accuracy} for the introduction and analyses of evaluating GEC systems by accuracy.

\begin{table*}[!tbp]
\centering
\scalebox{0.68}{
\begin{tabular}{lcllllllllllll}
\toprule
\multicolumn{1}{c}{\multirow{2}{*}{\textbf{Metric}}} & &
\multicolumn{2}{c}{\textbf{CoNLL-2014}} &
\multicolumn{2}{c}{\textbf{BN-10GEC}} &
\multicolumn{2}{c}{\textbf{E-Minimal}} & 
\multicolumn{2}{c}{\textbf{E-Fluency}} &
\multicolumn{2}{c}{\textbf{NE-Minimal}} &
\multicolumn{2}{c}{\textbf{NE-Fluency}} \\
\cmidrule(lr){3-4} \cmidrule(lr){5-6} \cmidrule(lr){7-8} \cmidrule(lr){9-10} \cmidrule(lr){11-12} \cmidrule(lr){13-14}

& & \multicolumn{1}{c}{\textbf{EW}} & \multicolumn{1}{c}{\textbf{TS}}
& \multicolumn{1}{c}{\textbf{EW}} & \multicolumn{1}{c}{\textbf{TS}}
& \multicolumn{1}{c}{\textbf{EW}} & \multicolumn{1}{c}{\textbf{TS}}
& \multicolumn{1}{c}{\textbf{EW}} & \multicolumn{1}{c}{\textbf{TS}}
& \multicolumn{1}{c}{\textbf{EW}} & \multicolumn{1}{c}{\textbf{TS}}
& \multicolumn{1}{c}{\textbf{EW}} & \multicolumn{1}{c}{\textbf{TS}} \\

\midrule

\multirow{2}{*}{\textbf{M$^2$}}
& $\gamma$ & 0.623 & 0.672 & 0.547 & 0.610 & \underline{0.597} & \underline{0.650} & 0.590 & 0.659 & 0.575 & 0.634 & 0.582 & 0.649 \\
& $\rho$   & 0.687 & 0.720 & 0.648 & 0.692 & 0.654 & 0.703 & 0.654 & 0.709 & 0.577 & 0.648 & 0.648 & 0.703 \\

\hdashline

\multirow{2}{*}{\textbf{GLEU}}
& $\gamma$ & \bf{0.701} & \bf{0.750} & \bf{0.678} & \bf{0.761} & 0.533 & 0.513 & \bf{0.693} & \bf{0.771} & -0.044 & -0.113 & \bf{0.674} & \bf{0.767} \\
& $\rho$   & 0.467 & 0.555 & \bf{0.754} & \underline{0.806} & 0.577 & 0.511 & 0.710 & 0.757 & -0.005 & -0.055 & 0.725 & \bf{0.819} \\

\hdashline

\multirow{2}{*}{\textbf{ERRANT}}
& $\gamma$ & 0.642 & 0.688 & 0.586 & 0.644 & 0.578 & 0.631 & 0.594 & 0.663 & 0.585 & 0.637 & 0.597 & 0.659 \\
& $\rho$   & 0.659 & 0.698 & 0.637 & 0.698 & 0.742 & 0.786 & 0.720 & 0.775 & 0.747 & 0.797 & \underline{0.753} & 0.797 \\

\hdashline

\multirow{2}{*}{\textbf{PT-M$^2$}}
& $\gamma$ & \underline{0.693} & \underline{0.737} & \underline{0.650} & \underline{0.706} & \bf{0.626} & \bf{0.667} & \underline{0.621} & \underline{0.681} & \bf{0.630} & \bf{0.675} & \underline{0.620} & \underline{0.682} \\
& $\rho$   & \bf{0.758} & \bf{0.769} & 0.690 & \bf{0.824} & 0.709 & 0.736 & \bf{0.758} & \bf{0.802} & 0.736 & 0.758 & \bf{0.758} & \underline{0.802} \\

\hdashline



\multirow{2}{*}{\textbf{CLEME-dependent} (Ours) }
& $\gamma$ & 0.648 & 0.691 & 0.602 & 0.656 & 0.594 & 0.644 & 0.589 & 0.654 & 0.595 & 0.643 & 0.612 & 0.673 \\
& $\rho$   & \underline{0.709} & \underline{0.742} & \underline{0.692} & 0.747 & \bf{0.797} & \bf{0.813} & 0.714 & 0.775 & \underline{0.786} & \underline{0.835} & 0.720 & 0.791 \\

\hdashline

\multirow{2}{*}{\textbf{CLEME-independent} (Ours)}
& $\gamma$ & 0.649 & 0.691 & 0.609 & 0.659 & 0.593 & 0.643 & 0.587 & 0.653 & \underline{0.601} & \underline{0.647} & 0.611 & 0.672 \\
& $\rho$   & \underline{0.709} & 0.731 & \underline{0.692} & 0.747 & \underline{0.791} & \underline{0.802} & \underline{0.731} & \underline{0.791} & \bf{0.797} & \bf{0.841} & 0.714 & 0.786 \\





\hline\hline

\multirow{2}{*}{\textbf{SentM$^2$}}
& $\gamma$ & 0.871 & \underline{0.864} & 0.567 & 0.646 & 0.805$^\clubsuit$ & 0.836$^\clubsuit$ & 0.655 & 0.732 & 0.729$^\clubsuit$ & 0.785$^\clubsuit$ & 0.621 & 0.699 \\
& $\rho$   & 0.731 & 0.758 & 0.593 & 0.648 & \underline{0.806}$^\clubsuit$ & \bf{0.845}$^\clubsuit$ & 0.731 & 0.764 & 0.797$^\clubsuit$ & 0.846$^\clubsuit$ & 0.632 & 0.687 \\

\hdashline

\multirow{2}{*}{\textbf{SentGLEU}}
& $\gamma$ & 0.784 & 0.828 & 0.756 & 0.826 & 0.742$^\clubsuit$ & 0.773$^\clubsuit$ & 0.785 & 0.846 & 0.723$^\clubsuit$ & 0.762$^\clubsuit$ & 0.778 & 0.848 \\
& $\rho$   & 0.720 & \underline{0.775} & 0.769 & 0.824 & 0.764$^\clubsuit$ & 0.797$^\clubsuit$ & 0.791 & \underline{0.846} & 0.764$^\clubsuit$ & 0.830$^\clubsuit$ & 0.768 & \underline{0.846} \\

\hdashline

\multirow{2}{*}{\textbf{SentERRANT} }
& $\gamma$ & 0.870 & 0.846 & \underline{0.885} & \underline{0.896} & 0.768$^\clubsuit$ & 0.803$^\clubsuit$ & 0.806 & 0.732 & 0.710$^\clubsuit$ & 0.765$^\clubsuit$ & 0.793 & 0.847 \\
& $\rho$   & 0.742 & 0.747 & \underline{0.786} & \underline{0.830} & 0.775$^\clubsuit$ & \underline{0.819}$^\clubsuit$ & \underline{0.813} & 0.764 & 0.780$^\clubsuit$ & 0.841$^\clubsuit$ & \bf{0.830} & \bf{0.857} \\

\hdashline

\multirow{2}{*}{\textbf{SentPT-M$^2$}}
& $\gamma$ & \bf{0.949} & \bf{0.938} & 0.602$^\clubsuit$ & 0.682$^\clubsuit$ & \bf{0.831}$^\clubsuit$ & \underline{0.855}$^\clubsuit$ & 0.689 & 0.763 & \underline{0.770}$^\clubsuit$ & \underline{0.822}$^\clubsuit$ & 0.648 & 0.725 \\
& $\rho$   & \bf{0.907} & \bf{0.874} & 0.626$^\clubsuit$ & 0.670$^\clubsuit$ & \bf{0.808}$^\clubsuit$ & \underline{0.819}$^\clubsuit$ & 0.797 & 0.841 & \underline{0.813}$^\clubsuit$ & \underline{0.857}$^\clubsuit$ & 0.742 & 0.786 \\

\hdashline



\multirow{2}{*}{\textbf{SentCLEME-dependent} (Ours)}
& $\gamma$ & \underline{0.876} & 0.844 & \bf{0.915} & \bf{0.913} & 0.806$^\clubsuit$ & 0.838$^\clubsuit$ & \bf{0.849} & \bf{0.886} & 0.742$^\clubsuit$ & 0.795$^\clubsuit$ & \bf{0.876} & \bf{0.921} \\
& $\rho$   & \underline{0.824} & \underline{0.808} & \bf{0.835} & \bf{0.874} & 0.775$^\clubsuit$ & \underline{0.819}$^\clubsuit$ & \bf{0.824} & \bf{0.863} & 0.797$^\clubsuit$ & 0.846$^\clubsuit$ & \underline{0.791} & \underline{0.846} \\

\hdashline

\multirow{2}{*}{\textbf{SentCLEME-independent} (Ours)}
& $\gamma$ & 0.868 & 0.857 & 0.855$^\clubsuit$ & 0.876$^\clubsuit$ & \underline{0.821}$^\clubsuit$ & \bf{0.856}$^\clubsuit$ & \underline{0.841} & \underline{0.877} & \bf{0.782}$^\clubsuit$ & \bf{0.831}$^\clubsuit$ & \underline{0.852} & \underline{0.896} \\
& $\rho$   & 0.725 & 0.758 & 0.659$^\clubsuit$ & 0.714$^\clubsuit$ & 0.775$^\clubsuit$ & \underline{0.819}$^\clubsuit$ & 0.808 & \underline{0.846} & \bf{0.819}$^\clubsuit$ & \bf{0.874}$^\clubsuit$ & 0.762 & 0.825 \\





\bottomrule
\end{tabular}}
\caption{\label{tab:exp-main}
Overview of correlations between mainstream GEC metrics and human judgments. We highlight the \textbf{highest} score in bold and the \underline{second-highest} score with underlines. SN-8GEC consists of four reference sets, i.e., E-Minimal, E-Fluency, NE-Minimal and NE-Fluency. $\clubsuit$ We remove unchanged reference sentences for higher correlations due to low-quality annotations. Otherwise, negative correlations are possible.}
\end{table*}

\paragraph{\textbf{Results}.}
Table~\ref{tab:exp-main} reports the correlations between reference-based metrics and human judgments.
For the corpus-level metrics, GLEU achieves the highest correlations on BN-10GEC and NE-fluency reference sets.
However, GLEU suffers from negative correlations on NE-Minimal, which is caused by low-quality annotations~\footnote{
The phenomenon exists on all sentence-level metrics. 
We remove unchanged references from some reference sets to avoid it.
} of NE-Minimal, indicating that GLEU may not be a robust metric, consistent with the findings of~\cite{sakaguchi-etal-2016-reassessing}.
ERRANT performs slightly better than M$^2$ on most reference sets, while PT-M$^2$ is a strong corpus-level metric, which achieves the highest or comparable correlations on all reference sets at the cost of more than 10$\times$ running time than other reference-based metrics.
Our proposed CLEME-dependent and CLEME-independent make better use of consistent chunk boundaries, thus performing slightly better than M$^2$ and ERRANT on most reference sets.
Notably, CLEME-independent achieves comparable performance to CLEME-dependent, showing the effectiveness of computing F$_{0.5}$ scores based on the correction independence assumption.

The majority of the sentence-level metrics outperform their corpus-level counterparts because they weigh samples equally, which is in line with the bias of human annotation.
Despite the strong performance of PT-M$^2$, SentPT-M$^2$ achieves lower correlations on BN-10GEC, E-Fluency and NE-Fluency compared to other sentence-level metrics.
It suggests that scoring edits using pre-trained language models may not generalize well to unseen reference sets for sentence-level metrics.
Our approach aligns better with human judgments than existing reference-based metrics for most reference sets.
Specifically, SentCLEME-dependent performs best on BN-10GEC and NE-Fluency, and performs on a par with the best metric on E-Fluency, indicating it is more suitable for fluent reference sets.
This phenomenon aligns with our intuition since fluent editing is more likely to follow the correction dependence assumption.
In contrast, SentCLEME-independent achieves higher correlations on E-Minimal and NE-Minimal, as we would expect from minimal editing that is more likely to follow the correction independence assumption.
These results suggests that reference sets may have a preference towards one of the correction assumptions.
Additionally, our approach achieves higher correlations on (N)E-Fluency rather than (N)E-Minimal, while SentM$^2$ and SentERRANT perform worse on E-Fluency than E-Minimal.
This is because CLEME evaluates GEC systems using longer chunks rather than scrappy edits, which could better reflect whether a grammatical error is fluently corrected.
Overall, our approach achieves higher or comparable correlations on sentence-level than existing reference-based methods.



\subsection{Human Evaluation}~\label{subsec:human_evaluation}
Experiments have shown the effectiveness of evaluating GEC systems based on the correction independence assumption.
In this section, we aim to demonstrate whether the correction independence assumption makes sense for humans.
We define the correction independence of a pair of chunks as the irrelevance of the correction of one chunk to the correction of the other.
A simple case is presented in Appendix~\ref{app:correction-independence}.
To evaluate this assumption, we conduct human evaluation experiments on 1,000 sentences randomly sampled from BN-10GEC~\cite{bryant-ng-2015-far}.
Three annotators were asked to judge whether a pair of chunks is correction-independent.

Table~\ref{tab:correction-independence} reports the ratio of correction independence and Cohen's-$\kappa$~\cite{cohen1960coefficient} inter-annotator agreement (IAA) across the three annotators.
Results show that more than 90\% pairs of chunks are correction-independent for all the annotators, indicating that it is reasonable to evaluate GEC systems based on the correction independence assumption.
Moreover, considering the subjectivity of GEC task, the IAA statistics show that it is relatively easy to judge whether a pair of chunks is correction-independent, compared with the previous study~\cite{bryant-ng-2015-far}~\footnote{
\citet{bryant-ng-2015-far} attempted to compute IAA at the sentence level.
Three raters were asked simply to decide whether 200 sentences were correct or not.
The authors reported IAA of just 0.16, 0.4 and 0.23.
}.

\begin{figure}[tbp]
\centering
\begin{tikzpicture}[scale=0.62]
\begin{axis}[
    legend style={
        text=black,
        font=\small,
        opacity=1.0,
        at={(1.01,1.00)},
        anchor=north west,
        legend cell align={left},
    },
    xlabel={Number of References},
    ylabel={Pearson Correlation $\gamma$},
    ylabel style = {yshift=-5pt},
    xmin=1, xmax=10,
    ymin=0.5, ymax=1.0,
    symbolic x coords={1,2,3,4,5,6,7,8,9,10},
    xtick=data,
    ytick={0.6,0.8,1.0},
    yticklabels={0.6,0.8,1.0},
    ymajorgrids=true,
    xmajorgrids=true,
    grid style=dashed,
]

\addplot[
    color=orange!80,dashed,thick,
    mark=square*,
    mark options={solid,mark size=2pt,rotate=45},
]
coordinates {
    (1,0.697)(2,0.660)(3,0.659)(4,0.672)(5,0.672)(6,0.676)(7,0.677)(8,0.680)(9,0.673)(10,0.682)
};
\addlegendentry{GLEU}

\addplot[
    color=black!80,thick,
    mark=triangle*,
    mark options={solid,mark size=2pt,rotate=90},
]
coordinates {
    (1,0.687)(2,0.750)(3,0.751)(4,0.757)(5,0.753)(6,0.755)(7,0.766)(8,0.764)(9,0.756)(10,0.756)
};
\addlegendentry{SentGLEU}

\addplot[
    color=magenta!80,dashed,thick,
    mark=halfcircle*,
    mark options={solid,mark size=2pt},
]
coordinates {
    (1,0.602)(2,0.593)(3,0.592)(4,0.591)(5,0.593)(6,0.589)(7,0.587)(8,0.585)(9,0.587)(10,0.586)
};
\addlegendentry{ERRANT}

\addplot[
    color=violet!80,thick,
    mark=*,
    mark options={solid,mark size=2pt},
]
coordinates {
    (1,0.711)(2,0.779)(3,0.800)(4,0.824)(5,0.836)(6,0.846)(7,0.875)(8,0.879)(9,0.878)(10,0.885)
};
\addlegendentry{SentERRANT}

\addplot[
    color=midnightblue!80,dashed,thick,
    mark=square*,
    mark options={solid,mark size=2pt}
]
coordinates {
    (1,0.596)(2,0.593)(3,0.599)(4,0.598)(5,0.602)(6,0.600)(7,0.598)(8,0.599)(9,0.602)(10,0.602)
};
\addlegendentry{CLEME-dependent}

\addplot[
    color=red!80,dashed,thick,
    mark=star,
    mark options={solid,mark size=2pt}
]
coordinates {
    (1,0.596)(2,0.593)(3,0.600)(4,0.600)(5,0.606)(6,0.602)(7,0.602)(8,0.603)(9,0.607)(10,0.609)
};
\addlegendentry{CLEME-independent}

\addplot[
    color=cyan!80,thick,
    mark=triangle*,
    mark options={solid,mark size=2pt}
]
coordinates {
    (1,0.721)(2,0.801)(3,0.829)(4,0.852)(5,0.871)(6,0.877)(7,0.908)(8,0.910)(9,0.910)(10,0.915)
};
\addlegendentry{SentCLEME-dependent}

\addplot[
    color=gray!80,thick,
    mark=*,
    mark options={solid,mark size=2pt},
]
coordinates {
    (1,0.648)(2,0.709)(3,0.751)(4,0.777)(5,0.802)(6,0.819)(7,0.829)(8,0.842)(9,0.850)(10,0.855)
};
\addlegendentry{SentCLEME-independent}
\end{axis}
\end{tikzpicture}
\caption{
Effect of references number on BN-10GEC.
}\label{fig:ablation_reference}
\end{figure}
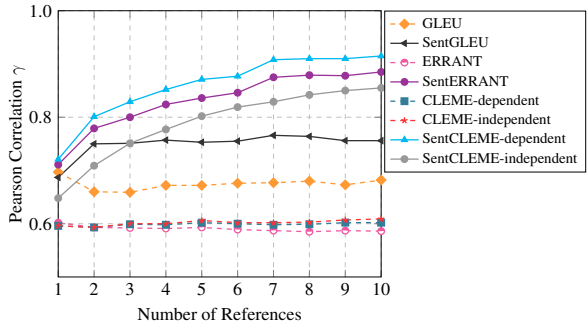

\begin{table}[tbp!]
\centering
\scalebox{0.74}{
\begin{tabular}{lc}
\toprule
\bf{Annotator} & \multicolumn{1}{c}{\bf{Ratio of Correction Independence}} \\
\midrule

A$_1$ & 90.85\% \\
A$_2$ & 93.55\% \\
A$_3$ & 91.14\% \\

\midrule
\bf{Annotator} & \multicolumn{1}{c}{\bf{Cohen's-$\kappa$}} \\
\midrule

A$_1$  v.s. A$_2$ & 38.66\% \\
A$_1$  v.s. A$_3$ & 43.10\% \\
A$_2$  v.s. A$_3$ & 39.34\% \\

\bottomrule
\end{tabular}}
\caption{\label{tab:correction-independence}
A comparison of correction independence annotations across three annotators.}
\end{table}

\section{Analysis}
\subsection{False Negative}

We observe that the number of false negatives (FNs) identified by CLEME is significantly lower than that of ERRANT.
This difference can be attributed to the distinct definitions used by each system.
While ERRANT considers FNs as edits in the reference that do not match those made in the hypothesis, CLEME identifies FNs as corrected/dummy chunks in the reference that do not match the chunks in the hypothesis.
We argue the definition of ERRANT is problematic, as it tends to overestimate FN counts in grammatical error correction (GEC) systems, which is evident from the examples presented in Table~\ref{tab:case-fn}.
On the other hand, CLEME's definition also includes true negatives (TNs), making it possible to calculate accuracy.



\begin{figure}[tbp!]
\centering
\begin{tikzpicture}[scale=0.62]
\begin{axis}[
    legend style={
        text=black,
        font=\small,
        opacity=1.0,
        at={(1.00,0.6)},
        anchor=north east,
        legend cell align={left},
    },
    xlabel={Scale Factor $\alpha$},
    xmin=2,   xmax=10,
    ymin=0.5, ymax=1.0,
    symbolic x coords={2,4,6,8,10},
    xtick=data,
    ytick={0.6,0.8,1.0},
    yticklabels={0.6,0.8,1.0},
    ymajorgrids=true,
    xmajorgrids=true,
    grid style=dashed,
]
\addplot[
    color=midnightblue!80,dashed,thick,
    mark=square*,
    mark options={solid,mark size=2pt}
]
coordinates {
    (2,0.602)(4,0.601)(6,0.600)(8,0.600)(10,0.600)
};
\addlegendentry{CLEME-dependent}

\addplot[
    color=red!80,dashed,thick,
    mark=star,
    mark options={solid,mark size=2pt}
]
coordinates {
    (2,0.604)(4,0.603)(6,0.602)(8,0.602)(10,0.602)
};
\addlegendentry{CLEME-independent}

\addplot[
    color=cyan!80,thick,
    mark=triangle*,
    mark options={solid,mark size=2pt}
]
coordinates {
    (2,0.915)(4,0.915)(6,0.915)(8,0.915)(10,0.915)
};
\addlegendentry{SentCLEME-dependent}

\addplot[
    color=gray!80,thick,
    mark=*,
    mark options={solid,mark size=2pt},
]
coordinates {
    (2,0.856)(4,0.855)(6,0.855)(8,0.855)(10,0.855)
};
\addlegendentry{SentCLEME-independent}
\end{axis}
\end{tikzpicture}
\caption{
Effect of scale factors on BN-10GEC. 
}\label{fig:ablation_parameter}
\end{figure}
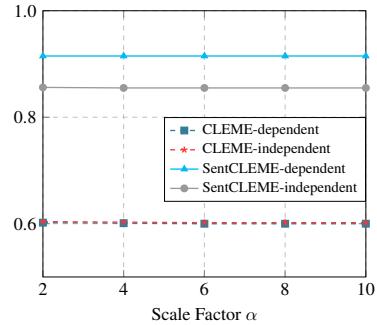

\begin{table}[tbp!]
\centering
\scalebox{0.64}{
\begin{tabular}{llcc}
\toprule
\bf{} & \multicolumn{1}{c}{\bf{Text}}
& \multicolumn{1}{c}{\bf{FP}} 
& \multicolumn{1}{c}{\bf{FN}} \\
\midrule

\bf{Source} & It has improved compared \textcolor{red}{for} the last century. \\

\bf{Hyp.} & It has improved compared \textcolor{red}{between} the last century. \\

\hline
\multicolumn{4}{c}{\bf{ERRANT}} \\
\hline

\bf{Ref. 1} & It has improved compared \textcolor{red}{to} the century. & 1 & 1 \\

\bf{Ref. 2} & It has improved compared \textcolor{red}{with} the century. & 1 & 1 \\

\hline
\multicolumn{4}{c}{\bf{CLEME}} \\
\hline

\bf{Ref. 1} & It has improved compared \textcolor{red}{to} the century. & 1 & 0 \\

\bf{Ref. 2} & It has improved compared \textcolor{red}{with} the century. & 1 & 0 \\

\bottomrule
\end{tabular}}
\caption{\label{tab:case-fn}
Cases of ERRANT and CLEME. ERRANT gives FP=1 and FN=1 since the hypothesis does not match one of the edits of references. CLEME gives only FP=1 since the hypothesis tries to correct the error.}
\end{table}
\begin{table*}[thbp!]
\centering
\scalebox{0.74}{
\begin{tabular}{lcccccc}
\toprule

& \bf{Chunk 1}
& \textcolor{red}{\bf{Chunk 2}}
& \bf{Chunk 3}
& \textcolor{red}{\bf{Chunk 4}}
& \bf{Chunk 5} 
& \textcolor{red}{\bf{Chunk 6}} \\

\hline

Source  &  On the other hand  ,  if there are  &  ways can  &  help us to control  &   or  &  cure the disease , we can  &  going .  \\

Hyp.  &  On the other hand , if there are  &  ways that can  &  help us to control  &  and  &  cure the disease , we can  &  go .  \\

Ref. 1  &  On the other hand , if there are  &  ways that can  &  help us to control  &  or  &  cure the disease , we can  &  go .  \\

Ref. 2  &  On the other hand , if there are  &  things that can  &  help us to control  &  and  &  cure the disease , we can  &  go .  \\

\bottomrule
\end{tabular}}
\end{table*}

\begin{table*}[thbp!]
\centering
\scalebox{0.75}{

\begin{tabular}{lccccc}
\toprule

& \bf{Chunk 1}
& \textcolor{red}{\bf{Chunk 2}}
& \bf{Chunk 3}
& \textcolor{red}{\bf{Chunk 4}}
& \bf{Chunk 5} \\

\hline

Source  &  On  &    &  one hand , we do not want this potential danger  &  causing firghtenning affects in  &  our lives . \\

Hyp.  &  On  &    &  one hand , we do not want this potential danger  &  causing frightening affects in  &  our lives .  \\

Ref. 1  &  On  &    &  one hand , we do not want this potential danger  & 
 having frightening effects in  &  our lives .  \\

Ref. 2  &  On  &  the  &  one hand , we do not want this potential danger  & 
 to have frightening effects on  &  our lives .  \\

\bottomrule
\end{tabular}}
\end{table*}

\begin{table*}[thbp!]
\centering
\scalebox{0.75}{
\begin{tabular}{lcccc}
\toprule

& \textcolor{red}{\bf{Chunk 1}}
& \bf{Chunk 2}
& \textcolor{red}{\bf{Chunk 3}}
& \bf{Chunk 4} \\

\hline

Source  &  Especially for the  &  young people  &  without marrige , if he/she is  &  known to have some genetic risk .   \\

Hyp.  &  Especially for the  &  young people  &  without marriage , if the latter is  &  known to have some genetic risk .  \\

Ref. 1  &  Especially for unmarried  &  young people  &  marrige who are  &  known to have some genetic risk .  \\

Ref. 2  &  This is especially the case for  &  young people  &  who are unmarried , if he/she is  &  known to have some genetic risk .  \\

\bottomrule
\end{tabular}}
\caption{
Cases of chunk partition.
These tables are automatically generated by CLEME.
More cases from multiple datasets and language are provided in Appendix~\ref{app:more-cases}.
}
\label{tab:case-study}
\end{table*}

\subsection{Ablation Study}
We present ablation analyses of our approaches on BN-10GEC - we have similar findings on other reference sets.
We report Pearson correlations $\gamma$ using Expected Wins ranking.
The trend is similar for Spearson correlations and TrueSkill ranking.

\paragraph{\textbf{Number of References}.}
Since CLEME is designed for multi-reference evaluation, it degrades to conventional reference-based metrics such as M${^2}$ and ERRANT when only one reference is available.
Here we demonstrate how correlations change against an increasing number of available references.
The results reported in Figure~\ref{fig:ablation_reference} indicate that the correlations of corpus-level metrics do not change significantly with the increasing number of available references.
However, except for SentGLEU, correlations of sentence-level metric are consistently higher than corpus-level metrics, and steadily increase with more references.
Therefore, we recommend evaluating GEC systems using sentence-level metrics rather than corpus-level metrics for the multi-reference evaluation setting.


\paragraph{\textbf{Parameter Sensitivity Analysis}.}
The scale factors introduced in Section~\ref{subsec:length-weighting} dictate how much the weights of chunks change with their length.
We report the corrections for various scale factors, as shown in Figure~\ref{fig:ablation_parameter}.
The results demonstrate that CLEME is resilient to hyperparameter selection.

\subsection{Case Study}
Table~\ref{tab:case-study} presents additional examples of CLEME.
In the top group, chunk 2 and chunk 4 of the hypothesis respectively match those of Ref. 1 and Ref. 2.
In this case, CLEME-dependent gives TP=1 and FP=1, while CLEME-independent gives TP=2.
In the second group, the hypothesis exactly corrects the ungrammatical word \textit{firghtenning} in chunk 4.
However, it cannot be rewarded since the entire chunk is not corrected.
In the bottom group, two given references have made extensive modifications, with an unchanged chunk \textit{young people}.
Evaluating hypotheses in terms of chunks is generally more challenging than fragmented edits, but it provides a more comprehensive diagnosis.

Even though there are larger grammatical errors spanning a significant portion of a sentence, CLEME would not necessarily \textit{collapse}, i.e., producing one single correction chunk spanning the entire sentence.
If collapse happens, the quality of the reference set should be checked first.
This is because that collapse happens only if the input sentences of chunk partition are completely different, resulting in a trivial chunk partition result, which is an extreme case that has not been observed in our experiments.

\section{Related Work}
\subsection{Reference-based Metrics}

\begin{table}[t]
\centering
\resizebox{\linewidth}{!}{
\begin{tabular}{lccc}
\toprule
\textbf{Reference-based Metrics} &\bf Granularity &\bf Score &\bf Deterministic\\
\midrule

M$^2$~\cite{dahlmeier-ng-2012-better} & Phrase-level Edit & F$_\beta$ & $\bullet$ \\

GLEU~\cite{napoles-etal-2015-ground}  & N-gram & Weighted Precision & $\circ$ \\


ERRANT~\cite{bryant-etal-2017-automatic}& Phrase-level Edit & F$_\beta$ & $\bullet$ \\

\hdashline

CLEME (Ours) & Chunk-level Edit & F$_\beta$ & $\bullet$ \\

\bottomrule
\end{tabular}}
\caption{\label{tab:comparison}
A comparison of mainstream reference-based GEC metrics.
GLEU is indeterministic since it involves sampling operation.
}
\end{table}

Reference-based metrics score GEC systems under the guidance of manually written references.
M$^2$ scorer~\cite{dahlmeier-ng-2012-better} determines an optimal edit sequence between a source sentence and a system hypothesis that achieves the highest overlap with the gold-standard annotation.
The performance of each system is then represented using the F$_{0.5}$ score.
However, optimality in terms of overlap does not guarantee optimality in GEC evaluation.
~\citet{bryant-etal-2017-automatic} showed that M$^2$ scorer exploits its dynamic edit boundary prediction to artificially maximize true positives and minimize false positives, thus producing slightly inflated scores.
Therefore, ~\cite{bryant-etal-2017-automatic} proposed ERRANT, which improves edit extraction using a linguistically-enhanced alignment algorithm and merging rules, improving the alignment of tokens with similar linguistic properties.
Despite its effectiveness, ERRANT is language-dependent and bias still exists in multi-reference evaluation.
Inspired by BLEU~\cite{papineni2002bleu} in NMT,~\citet{napoles-etal-2015-ground} proposed GLEU, an n-gram based metric for GEC evaluation.
To remedy the shortcoming that F$_{0.5}$ is unable to differentiate a do-nothing system and a bad system unless TP $>0$, I-measure~\cite{felice-briscoe-2015-towards} generates an exact (global optimal) alignment using a three-way alignment algorithm and computes weighted accuracy to score GEC systems in terms of relative textual improvement.
The comparison of reference-based GEC metrics is shown in Table~\ref{tab:comparison}.

\subsection{Reference-less Metrics}\label{subsec:ref-less}
To overcome the limitation of references for GEC evaluation, recent works focus on scoring GEC systems without the help of references.
Inspired by quality estimation in the NMT community, \citet{napoles-etal-2016-theres} proposed three Grammaticality-Based Metrics (GBMs) given by an existing GEC system or a pretrained ridge regression model.
\citet{asano-etal-2017-reference} extended GBMs~\cite{napoles-etal-2016-theres} with other assessment criteria, including grammaticality, fluency and meaning preservation.
SOME~\cite{yoshimura-etal-2020-reference} is a reference-less metric consisting of sub-metrics that are optimized for manual evaluation, which combines three regression models trained on the constructed dataset.
Scribendi Score~\cite{islam-magnani-2021-end} evaluates a GEC system using a combination of language model perplexity and sorted token/Levenshtein distance ratios.
IMPARA~\cite{maeda-etal-2022-impara} comprises a quality estimator (QE) and similarity estimator (SE) based on BERT~\cite{devlin-etal-2019-bert}, which evaluates the quality of the GEC output and semantic similarity of two sentences respectively.
Although reference-less metrics~\citet{napoles-etal-2016-theres} can achieve high agreement with human judgments, they lack interpretability as metrics for GEC evaluation.
Essentially, reference-less metrics are equivalent to evaluating GEC systems using other trained GEC systems, which could pose latent risk.
Additionally, the efficiency of reference-less metrics is also critical if used for GEC benchmark.

\subsection{Meta Evaluation Methods}\label{subsec:meta-eval}
It is intractable to determine the best GEC metric.
A reasonable GEC metric should take into account multiple factors, including correlation with human judgments, interpretability and efficiency.
Inspired by WMT human evaluation campaigns~\cite{callison-burch-etal-2008-meta}, 13 system outputs (including the unchanged source) from the CoNLL-2014 shared task~\cite{ng-etal-2014-conll} were ranked based on human rankings collected by two ranking methods: Expected Wins (EW) and TrueSkill (TS)~\cite{grundkiewicz2015human}.
\citet{sakaguchi-etal-2016-reassessing} collected 8 ($2\times 2\times 2$) annotations (minimal and fluency, expert and non-expert, with two corrections each), revealing that GEC metrics work differently across reference sets.
\citet{napoles-etal-2019-enabling} explored how GEC metrics operate in new domains (formal and informal writing by native English speakers).
They constructed a multi-reference GEC test set called GMEG-Data and a new ensemble metric GMEG-Metric.

\section{Conclusion}
\label{sec:conclusion}
This paper proposes CLEME, a reference-based GEC metric that aim to provide unbiased F$_{0.5}$ scores for multi-reference evaluation.
We explore evaluating GEC systems based on either the correction dependence assumption or the correction independence assumption.
Several possible approaches can be suggested to further improve CLEME.
For example, developing 
(1) a GEC metric that adaptively combines dependent and independent assumptions, and 
(2) a weighting strategy by utilizing the knowledge of pre-trained model.
In the future, we would like to develop CLEME for all languages and admonstrate the effectiveness of CLEME on other languages.
It is also worthwhile to explore accuracy-based metrics.

\section*{Limitations}
Although CLEME can be extended to other languages, we have not tested its effectiveness in any language other than English.
Furthermore, all the reference sets used in our experiments are based on the CoNLL-2014 shared task, a second-language dataset.
To demonstrate the robustness of our approaches, further experiments on evaluation datasets with multiple text domains are required.
We believe that introducing the correction independence assumption perspective into GEC datasets of other languages and domains could lead to more in-depth analysis and exploration.


While recent PLM-based metrics have shown superior correlations compared to reference-based metrics, including ours on some reference sets, our approach enables the evaluation of GEC systems in an interpretable manner, which is a significant advantage over reference-less metrics.
We leave the exploration of incorporating the PLM's knowledge into CLEME for future work.

\section*{Ethics Statement}
In this paper, we verify the effectiveness of our proposed approach using CoNLL-2014, BN-10GEC, and SN-8GEC reference sets, all of which are from publicly available datasets or resources on legitimate websites without sensitive data involved.
All the baselines used in our experiments are also publicly available GEC metrics and we have cited the corresponding authors.
We confirm that all datasets and baselines used in our experiments are consistent with their intended use.

Additionally, we conduct human evaluation experiments to show the rationality of correction independence assumption.
To do so, three postgraduate students specializing in foreign linguistics and applied linguistics were employed as part-time annotators.
Each annotator could complete the entire annotation process within approximately 6 working hours.
All annotators were paid for their work, with an average salary of approximately \$5 per hour.

\section*{Acknowledgements}
This research is supported by National Natural Science Foundation of China (Grant No.62276154), Research  Center for Computer Network(Shenzhen)Ministry of Education, the Natural Science Foundation of Guangdong Province (Grant No.  2023A1515012914), Basic Research Fund of Shenzhen City (Grant No. JCYJ20210324120012033 and JSGG20210802154402007), the Major Key Project of PCL for Experiments and Applications (PCL2021A06), and Overseas Cooperation Research Fund of Tsinghua Shenzhen International
Graduate School (HW2021008).

\bibliography{anthology,custom}

\begin{thebibliography}{38}
\expandafter\ifx\csname natexlab\endcsname\relax\def\natexlab#1{#1}\fi

\bibitem[{Asano et~al.(2017)Asano, Mizumoto, and
  Inui}]{asano-etal-2017-reference}
Hiroki Asano, Tomoya Mizumoto, and Kentaro Inui. 2017.
\newblock \href {https://aclanthology.org/I17-2058} {Reference-based metrics
  can be replaced with reference-less metrics in evaluating grammatical error
  correction systems}.
\newblock In \emph{Proceedings of the Eighth International Joint Conference on
  Natural Language Processing (Volume 2: Short Papers)}, pages 343--348,
  Taipei, Taiwan. Asian Federation of Natural Language Processing.

\bibitem[{Bryant et~al.(2017)Bryant, Felice, and
  Briscoe}]{bryant-etal-2017-automatic}
Christopher Bryant, Mariano Felice, and Ted Briscoe. 2017.
\newblock \href {https://doi.org/10.18653/v1/P17-1074} {Automatic annotation
  and evaluation of error types for grammatical error correction}.
\newblock In \emph{Proceedings of the 55th Annual Meeting of the Association
  for Computational Linguistics (Volume 1: Long Papers)}, pages 793--805,
  Vancouver, Canada. Association for Computational Linguistics.

\bibitem[{Bryant and Ng(2015)}]{bryant-ng-2015-far}
Christopher Bryant and Hwee~Tou Ng. 2015.
\newblock \href {https://doi.org/10.3115/v1/P15-1068} {How far are we from
  fully automatic high quality grammatical error correction?}
\newblock In \emph{Proceedings of the 53rd Annual Meeting of the Association
  for Computational Linguistics and the 7th International Joint Conference on
  Natural Language Processing (Volume 1: Long Papers)}, pages 697--707,
  Beijing, China. Association for Computational Linguistics.

\bibitem[{Bryant et~al.(2022)Bryant, Yuan, Qorib, Cao, Ng, and
  Briscoe}]{bryant2022grammatical}
Christopher Bryant, Zheng Yuan, Muhammad~Reza Qorib, Hannan Cao, Hwee~Tou Ng,
  and Ted Briscoe. 2022.
\newblock Grammatical error correction: A survey of the state of the art.
\newblock \emph{arXiv preprint arXiv:2211.05166}.

\bibitem[{Callison-Burch et~al.(2008)Callison-Burch, Fordyce, Koehn, Monz, and
  Schroeder}]{callison-burch-etal-2008-meta}
Chris Callison-Burch, Cameron Fordyce, Philipp Koehn, Christof Monz, and Josh
  Schroeder. 2008.
\newblock \href {https://aclanthology.org/W08-0309} {Further meta-evaluation of
  machine translation}.
\newblock In \emph{Proceedings of the Third Workshop on Statistical Machine
  Translation}, pages 70--106, Columbus, Ohio. Association for Computational
  Linguistics.

\bibitem[{Chodorow et~al.(2012)Chodorow, Dickinson, Israel, and
  Tetreault}]{chodorow-etal-2012-problems}
Martin Chodorow, Markus Dickinson, Ross Israel, and Joel Tetreault. 2012.
\newblock \href {https://aclanthology.org/C12-1038} {Problems in evaluating
  grammatical error detection systems}.
\newblock In \emph{Proceedings of {COLING} 2012}, pages 611--628, Mumbai,
  India. The COLING 2012 Organizing Committee.

\bibitem[{Chollampatt and Ng(2018)}]{chollampatt-ng-2018-reassessment}
Shamil Chollampatt and Hwee~Tou Ng. 2018.
\newblock \href {https://aclanthology.org/C18-1231} {A reassessment of
  reference-based grammatical error correction metrics}.
\newblock In \emph{Proceedings of the 27th International Conference on
  Computational Linguistics}, pages 2730--2741, Santa Fe, New Mexico, USA.
  Association for Computational Linguistics.

\bibitem[{Choshen and Abend(2018)}]{choshen-abend-2018-automatic}
Leshem Choshen and Omri Abend. 2018.
\newblock \href {https://doi.org/10.18653/v1/P18-1127} {Automatic metric
  validation for grammatical error correction}.
\newblock In \emph{Proceedings of the 56th Annual Meeting of the Association
  for Computational Linguistics (Volume 1: Long Papers)}, pages 1372--1382,
  Melbourne, Australia. Association for Computational Linguistics.

\bibitem[{Cohen(1960)}]{cohen1960coefficient}
Jacob Cohen. 1960.
\newblock A coefficient of agreement for nominal scales.
\newblock \emph{Educational and psychological measurement}, 20(1):37--46.

\bibitem[{Dahlmeier and Ng(2012)}]{dahlmeier-ng-2012-better}
Daniel Dahlmeier and Hwee~Tou Ng. 2012.
\newblock \href {https://aclanthology.org/N12-1067} {Better evaluation for
  grammatical error correction}.
\newblock In \emph{Proceedings of the 2012 Conference of the North {A}merican
  Chapter of the Association for Computational Linguistics: Human Language
  Technologies}, pages 568--572, Montr{\'e}al, Canada. Association for
  Computational Linguistics.

\bibitem[{Deutsch et~al.(2022)Deutsch, Dror, and Roth}]{deutsch2022limitations}
Daniel Deutsch, Rotem Dror, and Dan Roth. 2022.
\newblock On the limitations of reference-free evaluations of generated text.
\newblock \emph{arXiv preprint arXiv:2210.12563}.

\bibitem[{Devlin et~al.(2019)Devlin, Chang, Lee, and
  Toutanova}]{devlin-etal-2019-bert}
Jacob Devlin, Ming-Wei Chang, Kenton Lee, and Kristina Toutanova. 2019.
\newblock \href {https://doi.org/10.18653/v1/N19-1423} {{BERT}: Pre-training of
  deep bidirectional transformers for language understanding}.
\newblock In \emph{Proceedings of the 2019 Conference of the North {A}merican
  Chapter of the Association for Computational Linguistics: Human Language
  Technologies, Volume 1 (Long and Short Papers)}, pages 4171--4186,
  Minneapolis, Minnesota. Association for Computational Linguistics.

\bibitem[{Felice and Briscoe(2015)}]{felice-briscoe-2015-towards}
Mariano Felice and Ted Briscoe. 2015.
\newblock \href {https://doi.org/10.3115/v1/N15-1060} {Towards a standard
  evaluation method for grammatical error detection and correction}.
\newblock In \emph{Proceedings of the 2015 Conference of the North {A}merican
  Chapter of the Association for Computational Linguistics: Human Language
  Technologies}, pages 578--587, Denver, Colorado. Association for
  Computational Linguistics.

\bibitem[{Gong et~al.(2022)Gong, Liu, Huang, and Zhang}]{gong2022revisiting}
Peiyuan Gong, Xuebo Liu, Heyan Huang, and Min Zhang. 2022.
\newblock Revisiting grammatical error correction evaluation and beyond.
\newblock \emph{arXiv preprint arXiv:2211.01635}.

\bibitem[{Gotou et~al.(2020)Gotou, Nagata, Mita, and
  Hanawa}]{gotou-etal-2020-taking}
Takumi Gotou, Ryo Nagata, Masato Mita, and Kazuaki Hanawa. 2020.
\newblock \href {https://doi.org/10.18653/v1/2020.coling-main.188} {Taking the
  correction difficulty into account in grammatical error correction
  evaluation}.
\newblock In \emph{Proceedings of the 28th International Conference on
  Computational Linguistics}, pages 2085--2095, Barcelona, Spain (Online).
  International Committee on Computational Linguistics.

\bibitem[{Grundkiewicz et~al.(2015)Grundkiewicz, Junczys-Dowmunt, and
  Gillian}]{grundkiewicz2015human}
Roman Grundkiewicz, Marcin Junczys-Dowmunt, and Edward Gillian. 2015.
\newblock Human evaluation of grammatical error correction systems.
\newblock In \emph{Proceedings of the 2015 Conference on Empirical Methods in
  Natural Language Processing}, pages 461--470.

\bibitem[{Islam and Magnani(2021)}]{islam-magnani-2021-end}
Md~Asadul Islam and Enrico Magnani. 2021.
\newblock \href {https://doi.org/10.18653/v1/2021.emnlp-main.239} {Is this the
  end of the gold standard? a straightforward reference-less grammatical error
  correction metric}.
\newblock In \emph{Proceedings of the 2021 Conference on Empirical Methods in
  Natural Language Processing}, pages 3009--3015, Online and Punta Cana,
  Dominican Republic. Association for Computational Linguistics.

\bibitem[{Kaneko et~al.(2022)Kaneko, Takase, Niwa, and
  Okazaki}]{kaneko-etal-2022-interpretability}
Masahiro Kaneko, Sho Takase, Ayana Niwa, and Naoaki Okazaki. 2022.
\newblock \href {https://doi.org/10.18653/v1/2022.acl-long.496}
  {Interpretability for language learners using example-based grammatical error
  correction}.
\newblock In \emph{Proceedings of the 60th Annual Meeting of the Association
  for Computational Linguistics (Volume 1: Long Papers)}, pages 7176--7187,
  Dublin, Ireland. Association for Computational Linguistics.

\bibitem[{Li et~al.(2023)Li, Huang, Ma, Jiang, Li, Zhou, Zheng, and
  Zhou}]{DBLP:journals/corr/abs-2307-09007}
Yinghui Li, Haojing Huang, Shirong Ma, Yong Jiang, Yangning Li, Feng Zhou,
  Hai{-}Tao Zheng, and Qingyu Zhou. 2023.
\newblock \href {https://doi.org/10.48550/arXiv.2307.09007} {On the
  (in)effectiveness of large language models for chinese text correction}.
\newblock \emph{CoRR}, abs/2307.09007.

\bibitem[{Li et~al.(2022{\natexlab{a}})Li, Ma, Zhou, Li, Yangning, Huang, Liu,
  Li, Cao, and Zheng}]{li-etal-2022-learning-dictionary}
Yinghui Li, Shirong Ma, Qingyu Zhou, Zhongli Li, Li~Yangning, Shulin Huang,
  Ruiyang Liu, Chao Li, Yunbo Cao, and Haitao Zheng. 2022{\natexlab{a}}.
\newblock \href {https://doi.org/10.18653/v1/2022.findings-emnlp.18} {Learning
  from the dictionary: Heterogeneous knowledge guided fine-tuning for {C}hinese
  spell checking}.
\newblock In \emph{Findings of the Association for Computational Linguistics:
  EMNLP 2022}, pages 238--249, Abu Dhabi, United Arab Emirates. Association for
  Computational Linguistics.

\bibitem[{Li et~al.(2022{\natexlab{b}})Li, Zhou, Li, Li, Liu, Sun, Wang, Li,
  Cao, and Zheng}]{li-etal-2022-past}
Yinghui Li, Qingyu Zhou, Yangning Li, Zhongli Li, Ruiyang Liu, Rongyi Sun,
  Zizhen Wang, Chao Li, Yunbo Cao, and Hai-Tao Zheng. 2022{\natexlab{b}}.
\newblock \href {https://doi.org/10.18653/v1/2022.findings-acl.252} {The past
  mistake is the future wisdom: Error-driven contrastive probability
  optimization for {C}hinese spell checking}.
\newblock In \emph{Findings of the Association for Computational Linguistics:
  ACL 2022}, pages 3202--3213, Dublin, Ireland. Association for Computational
  Linguistics.

\bibitem[{Li et~al.(2021)Li, Parnow, Utiyama, Sumita, and
  Zhao}]{li-etal-2021-miss}
Zuchao Li, Kevin Parnow, Masao Utiyama, Eiichiro Sumita, and Hai Zhao. 2021.
\newblock \href {https://doi.org/10.18653/v1/2021.emnlp-demo.1} {{M}i{SS}: An
  assistant for multi-style simultaneous translation}.
\newblock In \emph{Proceedings of the 2021 Conference on Empirical Methods in
  Natural Language Processing: System Demonstrations}, pages 1--10, Online and
  Punta Cana, Dominican Republic. Association for Computational Linguistics.

\bibitem[{Ma et~al.(2023)Ma, Li, Huang, Huang, Li, Zheng, and
  Shen}]{DBLP:journals/corr/abs-2306-17447}
Shirong Ma, Yinghui Li, Haojing Huang, Shulin Huang, Yangning Li, Hai{-}Tao
  Zheng, and Ying Shen. 2023.
\newblock \href {https://doi.org/10.48550/arXiv.2306.17447} {Progressive
  multi-task learning framework for chinese text error correction}.
\newblock \emph{CoRR}, abs/2306.17447.

\bibitem[{Ma et~al.(2022)Ma, Li, Sun, Zhou, Huang, Zhang, Yangning, Liu, Li,
  Cao, Zheng, and Shen}]{ma-etal-2022-linguistic}
Shirong Ma, Yinghui Li, Rongyi Sun, Qingyu Zhou, Shulin Huang, Ding Zhang,
  Li~Yangning, Ruiyang Liu, Zhongli Li, Yunbo Cao, Haitao Zheng, and Ying Shen.
  2022.
\newblock \href {https://doi.org/10.18653/v1/2022.findings-emnlp.40}
  {Linguistic rules-based corpus generation for native {C}hinese grammatical
  error correction}.
\newblock In \emph{Findings of the Association for Computational Linguistics:
  EMNLP 2022}, pages 576--589, Abu Dhabi, United Arab Emirates. Association for
  Computational Linguistics.

\bibitem[{Mach{\'a}{\v{c}}ek and Bojar(2013)}]{machacek-bojar-2013-results}
Matou{\v{s}} Mach{\'a}{\v{c}}ek and Ond{\v{r}}ej Bojar. 2013.
\newblock \href {https://aclanthology.org/W13-2202} {Results of the {WMT}13
  metrics shared task}.
\newblock In \emph{Proceedings of the Eighth Workshop on Statistical Machine
  Translation}, pages 45--51, Sofia, Bulgaria. Association for Computational
  Linguistics.

\bibitem[{Maeda et~al.(2022)Maeda, Kaneko, and
  Okazaki}]{maeda-etal-2022-impara}
Koki Maeda, Masahiro Kaneko, and Naoaki Okazaki. 2022.
\newblock \href {https://aclanthology.org/2022.coling-1.316} {{IMPARA}:
  Impact-based metric for {GEC} using parallel data}.
\newblock In \emph{Proceedings of the 29th International Conference on
  Computational Linguistics}, pages 3578--3588, Gyeongju, Republic of Korea.
  International Committee on Computational Linguistics.

\bibitem[{Napoles et~al.(2019)Napoles, N{\u{a}}dejde, and
  Tetreault}]{napoles-etal-2019-enabling}
Courtney Napoles, Maria N{\u{a}}dejde, and Joel Tetreault. 2019.
\newblock \href {https://doi.org/10.1162/tacl_a_00282} {Enabling robust
  grammatical error correction in new domains: Data sets, metrics, and
  analyses}.
\newblock \emph{Transactions of the Association for Computational Linguistics},
  7:551--566.

\bibitem[{Napoles et~al.(2015)Napoles, Sakaguchi, Post, and
  Tetreault}]{napoles-etal-2015-ground}
Courtney Napoles, Keisuke Sakaguchi, Matt Post, and Joel Tetreault. 2015.
\newblock \href {https://doi.org/10.3115/v1/P15-2097} {Ground truth for
  grammatical error correction metrics}.
\newblock In \emph{Proceedings of the 53rd Annual Meeting of the Association
  for Computational Linguistics and the 7th International Joint Conference on
  Natural Language Processing (Volume 2: Short Papers)}, pages 588--593,
  Beijing, China. Association for Computational Linguistics.

\bibitem[{Napoles et~al.(2016)Napoles, Sakaguchi, and
  Tetreault}]{napoles-etal-2016-theres}
Courtney Napoles, Keisuke Sakaguchi, and Joel Tetreault. 2016.
\newblock \href {https://doi.org/10.18653/v1/D16-1228} {There{'}s no
  comparison: Reference-less evaluation metrics in grammatical error
  correction}.
\newblock In \emph{Proceedings of the 2016 Conference on Empirical Methods in
  Natural Language Processing}, pages 2109--2115, Austin, Texas. Association
  for Computational Linguistics.

\bibitem[{Napoles et~al.(2017)Napoles, Sakaguchi, and
  Tetreault}]{napoles-etal-2017-jfleg}
Courtney Napoles, Keisuke Sakaguchi, and Joel Tetreault. 2017.
\newblock \href {https://aclanthology.org/E17-2037} {{JFLEG}: A fluency corpus
  and benchmark for grammatical error correction}.
\newblock In \emph{Proceedings of the 15th Conference of the {E}uropean Chapter
  of the Association for Computational Linguistics: Volume 2, Short Papers},
  pages 229--234, Valencia, Spain. Association for Computational Linguistics.

\bibitem[{Ng et~al.(2014)Ng, Wu, Briscoe, Hadiwinoto, Susanto, and
  Bryant}]{ng-etal-2014-conll}
Hwee~Tou Ng, Siew~Mei Wu, Ted Briscoe, Christian Hadiwinoto, Raymond~Hendy
  Susanto, and Christopher Bryant. 2014.
\newblock \href {https://doi.org/10.3115/v1/W14-1701} {The {C}o{NLL}-2014
  shared task on grammatical error correction}.
\newblock In \emph{Proceedings of the Eighteenth Conference on Computational
  Natural Language Learning: Shared Task}, pages 1--14, Baltimore, Maryland.
  Association for Computational Linguistics.

\bibitem[{Papineni et~al.(2002)Papineni, Roukos, Ward, and
  Zhu}]{papineni2002bleu}
Kishore Papineni, Salim Roukos, Todd Ward, and Wei-Jing Zhu. 2002.
\newblock Bleu: a method for automatic evaluation of machine translation.
\newblock In \emph{Proceedings of the 40th annual meeting of the Association
  for Computational Linguistics}, pages 311--318.

\bibitem[{Sakaguchi et~al.(2016)Sakaguchi, Napoles, Post, and
  Tetreault}]{sakaguchi-etal-2016-reassessing}
Keisuke Sakaguchi, Courtney Napoles, Matt Post, and Joel Tetreault. 2016.
\newblock \href {https://doi.org/10.1162/tacl_a_00091} {Reassessing the goals
  of grammatical error correction: Fluency instead of grammaticality}.
\newblock \emph{Transactions of the Association for Computational Linguistics},
  4:169--182.

\bibitem[{Sakaguchi et~al.(2014)Sakaguchi, Post, and
  Van~Durme}]{sakaguchi-etal-2014-efficient}
Keisuke Sakaguchi, Matt Post, and Benjamin Van~Durme. 2014.
\newblock \href {https://doi.org/10.3115/v1/W14-3301} {Efficient elicitation of
  annotations for human evaluation of machine translation}.
\newblock In \emph{Proceedings of the Ninth Workshop on Statistical Machine
  Translation}, pages 1--11, Baltimore, Maryland, USA. Association for
  Computational Linguistics.

\bibitem[{Ye et~al.(2022)Ye, Li, Ma, Xie, Wu, and
  Zheng}]{DBLP:journals/corr/abs-2210-12692}
Jingheng Ye, Yinghui Li, Shirong Ma, Rui Xie, Wei Wu, and Hai{-}Tao Zheng.
  2022.
\newblock \href {https://doi.org/10.48550/arXiv.2210.12692} {Focus is what you
  need for chinese grammatical error correction}.
\newblock \emph{CoRR}, abs/2210.12692.

\bibitem[{Yoshimura et~al.(2020)Yoshimura, Kaneko, Kajiwara, and
  Komachi}]{yoshimura-etal-2020-reference}
Ryoma Yoshimura, Masahiro Kaneko, Tomoyuki Kajiwara, and Mamoru Komachi. 2020.
\newblock \href {https://doi.org/10.18653/v1/2020.coling-main.573} {{SOME}:
  Reference-less sub-metrics optimized for manual evaluations of grammatical
  error correction}.
\newblock In \emph{Proceedings of the 28th International Conference on
  Computational Linguistics}, pages 6516--6522, Barcelona, Spain (Online).
  International Committee on Computational Linguistics.

\bibitem[{Zhang et~al.(2023)Zhang, Li, Zhou, Ma, Li, Cao, and Zheng}]{10095675}
Ding Zhang, Yinghui Li, Qingyu Zhou, Shirong Ma, Yangning Li, Yunbo Cao, and
  Hai-Tao Zheng. 2023.
\newblock \href {https://doi.org/10.1109/ICASSP49357.2023.10095675} {Contextual
  similarity is more valuable than character similarity: An empirical study for
  chinese spell checking}.
\newblock In \emph{ICASSP 2023 - 2023 IEEE International Conference on
  Acoustics, Speech and Signal Processing (ICASSP)}, pages 1--5.

\bibitem[{Zhang et~al.(2022)Zhang, Li, Bao, Li, Zhang, Li, Huang, and
  Zhang}]{zhang-etal-2022-mucgec}
Yue Zhang, Zhenghua Li, Zuyi Bao, Jiacheng Li, Bo~Zhang, Chen Li, Fei Huang,
  and Min Zhang. 2022.
\newblock \href {https://doi.org/10.18653/v1/2022.naacl-main.227} {{M}u{CGEC}:
  a multi-reference multi-source evaluation dataset for {C}hinese grammatical
  error correction}.
\newblock In \emph{Proceedings of the 2022 Conference of the North American
  Chapter of the Association for Computational Linguistics: Human Language
  Technologies}, pages 3118--3130, Seattle, United States. Association for
  Computational Linguistics.

\end{thebibliography}
\bibliographystyle{acl_natbib}

\appendix

\appendix







\begin{table*}[htbp!]
\centering
\scalebox{0.67}{
\begin{tabular}{lcccccc}
\toprule
\bf{Item} & \bf{CoNLL-2014} & \bf{BN-10GEC} & \bf{E-Minimal} & \bf{E-Fluency} & \bf{NE-Minimal} & \bf{NE-Fluency} \\
\midrule

\# Sents (Length) & 1,312 (23.0) & 1,312 (23.0) & 1,312 (23.0) & 1,312 (23.0) & 1,312 (23.0) & 1,312 (23.0) \\
\# Refs (Length) & 2,624 (22.8) & 13,120 (22.9) & 2,624 (23.2) & 2,624 (22.8) & 2,624 (23.0) & 2,624 (22.2) \\
\# Edits (Length) & 5,937 (1.0) & 36,677 (1.0) & 4,500 (1.0) & 8,373 (1.1) & 4,964 (0.9) & 11,033 (1.2) \\
\# Unchanged Chunks (Length) & 11,174 (4.8) & 93,496 (2.5) & 8,887 (6.3) & 12,823 (3.8) & 10,748 (5.1) & 14,086 (2.9) \\
\# Corrected/Dummy Chunks (Length) & 4,994 (1.3) & 26,948 (2.4) & 3,963 (1.2) & 6,305 (1.7) & 4,221 (1.2) & 6,892 (2.6) \\
ICC (Number) & 51.05\% (3,031) & 90.66\% (33,251) & 62.47\% (2,811) & 52.69\% (4,412) & 43.84\% (2,176) & 43.32\% (4,780) \\
IUC (Number) & 45.61\% (2,708) & 7.74\% (2,837) & 36.6\% (1,647) & 42.51\% (3,559) & 54.3\% (2,698) & 48.92\% (5,397) \\
CC (Number) & 3.34\% (198) & 1.61\% (589) & 0.93\% (42) & 4.8\% (402) & 1.81\% (90) & 7.76\% (856) \\

\bottomrule
\end{tabular}}
\caption{\label{tab:statistics-more-ref}
Statistics of CoNLL-2014~\cite{ng-etal-2014-conll}, BN-10GEC~\cite{bryant-ng-2015-far} and SN-8GEC~\cite{sakaguchi-etal-2016-reassessing} reference sets. We use ERRANT~\cite{bryant-etal-2017-automatic} for edit extraction.}
\end{table*}
\begin{table*}[htbp!]
\centering
\scalebox{0.64}{
\begin{tabular}{lcccc}
\toprule
\textbf{Configuration} 
& \textbf{CLEME-dependent(-acc)}
& \textbf{CLEME-independent(-acc)}
& \textbf{SentCLEME-dependent(-acc)}
& \textbf{SentCLEME-independent(-acc)} \\

\hline
Scale factor of TPs $\alpha_1$ & 2 & 2 & 10 & 10 \\
Scale factor of FPs $\alpha_2$ & 2 & 2 & 10 & 10 \\
Scale factor of FNs $\alpha_3$ & 2 & 2 & - & - \\
Scale factor of TNs $\alpha_4$ & - & - & - & - \\
Threshold of TPs & $(0.75,1.25)$ & $(0.75,1.25)$ & $(1.00,10.00)$ & $(2.50,10.00)$ \\
Threshold of FPs & $(0.75,1.25)$ & $(0.75,1.25)$ & $(0.25,10.00)$ & $(0.25,1.00)$ \\
Threshold of FNs & $(0.75,1.25)$ & $(0.75,1.25)$ & $(1.00,1.00)$ & $(1.00,1.00)$ \\
Threshold of TNs & $(1.00,1.00)$ & $(1.00,1.00)$ & $(1.00,1.00)$ & $(1.00,1.00)$ \\

\bottomrule
\end{tabular}}
\caption{Hyperparameter values used in our experiments.}
\label{tab:hyperparameters}
\end{table*}

\section{Statistics of Reference Sets}\label{subsec:statistics-ref}
Table~\ref{tab:statistics-more-ref} presents the statistics of all reference sets involved in our experiments, including In-Corrected-Chunk (ICC) ratio, Unchanged-Chunk (IUC) ratio and Cross-Chunk (CC) ratio.
It is worth noting that all reference sets exhibit a low CC ratio with varying ICC and IUC ratios, indicating the rationality and feasibility of evaluating GEC systems following the correction independence assumption.

\section{Hyperparameters}\label{subsec:hyperparameters}
The hyperparameters of our proposed CLEME consist of scale factors $\alpha$ and thresholds.
We tune the hyperparameters on CoNLL-2014 and keep them on the other reference sets to demonstrate the adaptability of our method.
The hyperparameters of CLEME are listed in Table~\ref{tab:hyperparameters}.

\section{Evaluate by Accuracy}\label{subsec:accuracy}
Conventional reference-based metrics such as MaxMatch (M$^2$) and ERRANT are unable to calculate accuracy because they do not define True Negatives (TNs)~\footnote{
An exception is I-measure~\cite{felice-briscoe-2015-towards}, which adopts an extended version of the Writer-Annotation-System evaluation scheme~\cite{chodorow-etal-2012-problems}.
}.
In order to implement the computation of accuracy, CLEME defines TNs as hypothesis unchanged chunks that match the chunks of references.
Similar to F$_{0.5}$, accuracy can be computed based on correction dependence or independence assumptions in both corpus- and sentence-level, resulting in four new variants:
1) \textbf{CLEME-dependent-acc}, 
2) \textbf{CLEME-independent-acc}, 
3) \textbf{SentCLEME-dependent-acc}, and 
4) \textbf{SentCLEME-independent-acc}.

The results of human correlations are reported in Table~\ref{tab:exp-summary}.
Accuracy-based metrics perform very differently at the corpus- and sentence-level, which is similar to the findings~\cite{napoles-etal-2016-theres,napoles-etal-2019-enabling}.
Surprisingly, two accuracy-based corpus-level metrics, i.e., CLEME-dependent-acc and CLEME-independent-acc, result in negative correlations on all reference sets.
However, their sentence-level variants, i.e., SentCLEME-dependent-acc and SentCLEME-independent-acc, perform well and achieve the highest correlations on some reference sets.
Regarding the disparity between the performance of accuracy-based metrics and F$_{0.5}$-based metrics at the sentence level, one notable difference is their stability or robustness on reference sets with varying numbers of references and annotation styles.
F$_{0.5}$-based metrics are more robust to different reference sets, where SentCLEME-(in)dependent achieves comparable correlations with the best metric on all reference sets.
However, the performance of accuracy-based metrics lags far behind other metrics on some reference sets (BN-10GEC, E-Minimal and NE-Fluency).
A deeper investigation into this phenomenon is needed to understand the instability of accuracy-based metrics.
We leave the exploration and further analysis of accuracy-based metric for future work.

\begin{table*}[!tbp]
\centering
\scalebox{0.62}{
\begin{tabular}{lcllllllllllll}
\toprule
\multicolumn{1}{c}{\multirow{2}{*}{\textbf{Metric}}} & &
\multicolumn{2}{c}{\textbf{CoNLL-2014}} &
\multicolumn{2}{c}{\textbf{BN-10GEC}} &
\multicolumn{2}{c}{\textbf{E-Minimal}} & 
\multicolumn{2}{c}{\textbf{E-Fluency}} &
\multicolumn{2}{c}{\textbf{NE-Minimal}} &
\multicolumn{2}{c}{\textbf{NE-Fluency}} \\
\cmidrule(lr){3-4} \cmidrule(lr){5-6} \cmidrule(lr){7-8} \cmidrule(lr){9-10} \cmidrule(lr){11-12} \cmidrule(lr){13-14}

& & \multicolumn{1}{c}{\textbf{EW}} & \multicolumn{1}{c}{\textbf{TS}}
& \multicolumn{1}{c}{\textbf{EW}} & \multicolumn{1}{c}{\textbf{TS}}
& \multicolumn{1}{c}{\textbf{EW}} & \multicolumn{1}{c}{\textbf{TS}}
& \multicolumn{1}{c}{\textbf{EW}} & \multicolumn{1}{c}{\textbf{TS}}
& \multicolumn{1}{c}{\textbf{EW}} & \multicolumn{1}{c}{\textbf{TS}}
& \multicolumn{1}{c}{\textbf{EW}} & \multicolumn{1}{c}{\textbf{TS}} \\

\midrule

\multirow{2}{*}{\textbf{M$^2$}}
& $\gamma$ & 0.623 & 0.672 & 0.547 & 0.610 & \bf{0.597} & \bf{0.650} & \underline{0.590} & 0.659 & 0.575 & 0.634 & 0.582 & 0.649 \\
& $\rho$   & \underline{0.687} & 0.720 & 0.648 & 0.692 & 0.654 & 0.703 & 0.654 & 0.709 & 0.577 & 0.648 & 0.648 & 0.703 \\

\hdashline

\multirow{2}{*}{\textbf{GLEU}}
& $\gamma$ & \bf{0.701} & \bf{0.750} & \bf{0.678} & \bf{0.761} & 0.533 & 0.513 & \bf{0.693} & \bf{0.771} & -0.044 & -0.113 & \bf{0.674} & \bf{0.767} \\
& $\rho$   & 0.467 & 0.555 & \bf{0.754} & \bf{0.806} & 0.577 & 0.511 & 0.710 & 0.757 & -0.005 & -0.055 & \underline{0.725} & \bf{0.819} \\

\hdashline

\multirow{2}{*}{\textbf{ERRANT}}
& $\gamma$ & 0.642 & 0.688 & 0.586 & 0.644 & 0.578 & 0.631 & 0.594 & \underline{0.663} & 0.585 & 0.637 & 0.597 & 0.659 \\
& $\rho$   & 0.659 & 0.698 & 0.637 & 0.698 & 0.742 & 0.786 & \underline{0.720} & \underline{0.775} & 0.747 & 0.797 & \bf{0.753} & \underline{0.797} \\

\hdashline

\multirow{2}{*}{\textbf{CLEME-dependent} (Ours) }
& $\gamma$ & 0.648 & \underline{0.691} & 0.602 & 0.656 & \underline{0.594} & \underline{0.644} & 0.589 & 0.654 & \underline{0.595} & \underline{0.643} & \underline{0.612} & \underline{0.673} \\
& $\rho$   & \bf{0.709} & \bf{0.742} & \underline{0.692} & \underline{0.747} & \bf{0.797} & \bf{0.813} & 0.714 & \underline{0.775} & \underline{0.786} & \underline{0.835} & 0.720 & 0.791 \\

\hdashline

\multirow{2}{*}{\textbf{CLEME-independent} (Ours)}
& $\gamma$ & \underline{0.649} & \underline{0.691} & \underline{0.609} & \underline{0.659} & 0.593 & 0.643 & 0.587 & 0.653 & \bf{0.601} & \bf{0.647} & 0.611 & 0.672 \\
& $\rho$   & \bf{0.709} & \underline{0.731} & \underline{0.692} & \underline{0.747} & \underline{0.791} & \underline{0.802} & \bf{0.731} & \bf{0.791} & \bf{0.797} & \bf{0.841} & 0.714 & 0.786 \\

\hdashline

\multirow{2}{*}{\textbf{CLEME-dependent-acc} (Ours)} 
& $\gamma$ & -0.261 & -0.342 & -0.288 & -0.371 & -0.222 & -0.313 & -0.216 & -0.302 & -0.370 & -0.453 & -0.430 & -0.513 \\
& $\rho$   & -0.407 & -0.478 & -0.445 & -0.516 & -0.335 & -0.423 & -0.347 & -0.437 & -0.429 & -0.516 & -0.473 & -0.555 \\

\hdashline

\multirow{2}{*}{\textbf{CLEME-independent-acc} (Ours)} 
& $\gamma$ & -0.175 & -0.262 & -0.206 & -0.284 & -0.195 & -0.283 & -0.105 & -0.189 & -0.335 & -0.420 & -0.328 & -0.412 \\
& $\rho$   & -0.176 & -0.264 & -0.341 & -0.418 & -0.291 & -0.379 & -0.132 & -0.231 & -0.429 & -0.516 & -0.451 & -0.522 \\ 

\hline\hline

\multirow{2}{*}{\textbf{SentM$^2$}}
& $\gamma$ & 0.871 & 0.864 & 0.567 & 0.646 & 0.805$^\clubsuit$ & 0.836$^\clubsuit$ & 0.655 & 0.732 & 0.729$^\clubsuit$ & 0.785$^\clubsuit$ & 0.621 & 0.699 \\
& $\rho$   & 0.731 & 0.758 & 0.593 & 0.648 & \underline{0.806}$^\clubsuit$ & \underline{0.845}$^\clubsuit$ & 0.731 & 0.764 & 0.797$^\clubsuit$ & 0.846$^\clubsuit$ & 0.632 & 0.687 \\

\hdashline

\multirow{2}{*}{\textbf{SentGLEU}}
& $\gamma$ & 0.784 & 0.828 & 0.756 & 0.826 & 0.742$^\clubsuit$ & 0.773$^\clubsuit$ & 0.785 & 0.846 & 0.723$^\clubsuit$ & 0.762$^\clubsuit$ & 0.778 & 0.848 \\
& $\rho$   & 0.720 & 0.775 & 0.769 & 0.824 & 0.764$^\clubsuit$ & 0.797$^\clubsuit$ & 0.791 & 0.846 & 0.764$^\clubsuit$ & 0.830$^\clubsuit$ & 0.768 & \underline{0.846} \\

\hdashline

\multirow{2}{*}{\textbf{SentERRANT} }
& $\gamma$ & 0.870 & 0.846 & \underline{0.885} & \underline{0.896} & 0.768$^\clubsuit$ & 0.803$^\clubsuit$ & 0.806 & 0.732 & 0.710$^\clubsuit$ & 0.765$^\clubsuit$ & 0.793 & 0.847 \\
& $\rho$   & 0.742 & 0.747 & \underline{0.786} & \underline{0.830} & 0.775$^\clubsuit$ & 0.819$^\clubsuit$ & 0.813 & 0.764 & 0.780$^\clubsuit$ & 0.841$^\clubsuit$ & \bf{0.830} & \bf{0.857} \\

\hdashline

\multirow{2}{*}{\textbf{SentCLEME-dependent} (Ours)}
& $\gamma$ & \underline{0.876} & 0.844 & \bf{0.915} & \bf{0.913} & 0.806$^\clubsuit$ & \underline{0.838}$^\clubsuit$ & \bf{0.849} & \underline{0.886} & 0.742$^\clubsuit$ & 0.795$^\clubsuit$ & \bf{0.876} & \bf{0.921} \\
& $\rho$   & \underline{0.824} & 0.808 & \bf{0.835} & \bf{0.874} & 0.775$^\clubsuit$ & 0.819$^\clubsuit$ & \underline{0.824} & \underline{0.863} & 0.797$^\clubsuit$ & 0.846$^\clubsuit$ & \underline{0.791} & \underline{0.846} \\

\hdashline

\multirow{2}{*}{\textbf{SentCLEME-independent} (Ours)}
& $\gamma$ & 0.868 & \underline{0.857} & 0.855$^\clubsuit$ & 0.876$^\clubsuit$ & \bf{0.821}$^\clubsuit$ & \bf{0.856}$^\clubsuit$ & \underline{0.841} & 0.877 & \underline{0.782}$^\clubsuit$ & \bf{0.831}$^\clubsuit$ & \underline{0.852} & \underline{0.896} \\
& $\rho$   & 0.725 & 0.758 & 0.659$^\clubsuit$ & 0.714$^\clubsuit$ & 0.775$^\clubsuit$ & 0.819$^\clubsuit$ & 0.808 & 0.846 & \underline{0.819}$^\clubsuit$ & \underline{0.874}$^\clubsuit$ & 0.762 & 0.825 \\

\hdashline

\multirow{2}{*}{\textbf{SentCLEME-dependent-acc} (Ours)}
& $\gamma$ & 0.828 & \underline{0.857} & 0.650 & 0.719 & \underline{0.808}$^\clubsuit$ & \underline{0.838}$^\clubsuit$ & 0.679 & 0.740 & 0.757$^\clubsuit$ & \underline{0.811}$^\clubsuit$ & 0.557 & 0.641 \\
& $\rho$   & 0.813 & \underline{0.841} & 0.682 & 0.740 & \bf{0.830}$^\clubsuit$ & \bf{0.852}$^\clubsuit$ & 0.731 & 0.786 & \bf{0.853}$^\clubsuit$ & \bf{0.894}$^\clubsuit$ & 0.655 & 0.702 \\

\hdashline

\multirow{2}{*}{\textbf{SentCLEME-independent-acc} (Ours)}
& $\gamma$ & \bf{0.900}$^\clubsuit$ & \bf{0.920}$^\clubsuit$ & 0.604$^\clubsuit$ & 0.555$^\clubsuit$ & 0.693$^\clubsuit$ & 0.637$^\clubsuit$ & 0.840$^\clubsuit$ & \bf{0.891}$^\clubsuit$ & \bf{0.791}$^\clubsuit$ & 0.763$^\clubsuit$ & 0.756$^\clubsuit$ & 0.822$^\clubsuit$ \\
& $\rho$   & \bf{0.830}$^\clubsuit$ & \bf{0.849}$^\clubsuit$ & 0.363$^\clubsuit$ & 0.303$^\clubsuit$ & 0.588$^\clubsuit$ & 0.544$^\clubsuit$ & \bf{0.857}$^\clubsuit$ & \bf{0.890}$^\clubsuit$ & 0.654$^\clubsuit$ & 0.626$^\clubsuit$ & 0.747$^\clubsuit$ & 0.819$^\clubsuit$ \\

\bottomrule
\end{tabular}}
\caption{\label{tab:exp-summary}
Overview of correlations between reference-based metrics and human judgments. We highlight the \textbf{highest} score in bold and the \underline{second-highest} score with underlines. $\clubsuit$ We remove unchanged reference sentences for higher correlations due to low-quality annotations. Otherwise, negative correlations are possible.}
\end{table*}
\begin{table*}[tbp!]
\centering
\scalebox{0.64}{
\begin{tabular}{llccccccccccccc}
\toprule
\bf{Metric} &
& \multicolumn{1}{c}{\bf{AMU}}
& \multicolumn{1}{c}{\bf{CAMB}}
& \multicolumn{1}{c}{\bf{CUUI}}
& \multicolumn{1}{c}{\bf{IITB}}
& \multicolumn{1}{c}{\bf{INPUT}}
& \multicolumn{1}{c}{\bf{IPN}}
& \multicolumn{1}{c}{\bf{NTHU}}
& \multicolumn{1}{c}{\bf{PKU}}
& \multicolumn{1}{c}{\bf{POST}}
& \multicolumn{1}{c}{\bf{RAC}}
& \multicolumn{1}{c}{\bf{SJTU}}
& \multicolumn{1}{c}{\bf{UFC}}
& \multicolumn{1}{c}{\bf{UMC}} \\
\hline

\multirow{6}{*}{\bf{ERRANT}} 
& \bf{TP} & 483 & 725 & 607 & 28 & 0 & 52 & 409 & 294 & 508 & 319 & 104 & 36 & 311 \\
& \bf{FP} & 795 & 1329 & 985 & 65 & 0 & 488 & 991 & 697 & 1152 & 794 & 261 & 14 & 774 \\
& \bf{FN} & 1934 & 1886 & 1946 & 2064 & 2070 & 2078 & 1976 & 2007 & 1985 & 2044 & 2036 & 2069 & 2020 \\
& \bf{P} & 37.79 & 35.30 & 38.13 & 30.11 & 100.0 & 9.63 & 29.21 & 29.67 & 30.60 & 28.66 & 28.49 & 72.00 & 28.66 \\
& \bf{R} & 19.98 & 27.77 & 23.78 & 1.34 & 0.00 & 2.44 & 17.15 & 12.78 & 20.38 & 13.50 & 4.86 & 1.71 & 13.34 \\
& \bf{F$_{0.5}$} & 32.08 & 33.48 & 34.02 & 5.68 & 0.00 & 6.06 & 25.61 & 23.46 & 27.81 & 23.40 & 14.44 & 7.81 & 23.31 \\

\midrule

\multirow{11}{*}{\bf{CLEME-dependent}}
& \bf{TP} & 314 & 482 & 379 & 17 & 0 & 33 & 266 & 195 & 333 & 203 & 69 & 24 & 213 \\
& \quad w/o LW & 382 & 588 & 471 & 22 & 0 & 39 & 330 & 246 & 412 & 254 & 85 & 32 & 216 \\
& \bf{FP} & 872 & 1392 & 1034 & 72 & 0 & 529 & 975 & 776 & 1246 & 782 & 292 & 19 & 844 \\
& \quad w/o LW & 815 & 1303 & 964 & 67 & 0 & 488 & 905 & 709 & 1144 & 782 & 272 & 18 & 788 \\
& \bf{FN} & 1182 & 987 & 1169 & 1564 & 1592 & 1445 & 1191 & 1259 & 1158 & 1278 & 1471 & 1583 & 1266 \\
& \quad w/o LW & 1345 & 1132 & 1333 & 1751 & 1782 & 1634 & 1366 & 1426 & 1333 & 1453 & 1657 & 1772 & 1439 \\
& \bf{TN} & 6312 & 6347 & 6245 & 6313 & 6308 & 6412 & 6295 & 6314 & 6449 & 6310 & 6324 & 6280 & 6377 \\

& \bf{P} & 26.45 & 25.74 & 26.81 & 19.29 & 100.0 & 5.85 & 21.42 & 20.06 & 21.07 & 20.60 & 19.02 & 56.40 & 20.14 \\
& \bf{R} & 20.97 & 32.84 & 24.48 & 1.09 & 0.00 & 2.22 & 18.23 & 13.39 & 22.31 & 13.71 & 4.45 & 1.52 & 14.40 \\
& \bf{F$_{0.5}$} & 25.14 & 26.90 & 26.31 & 4.45 & 0.00 & 4.41 & 20.69 & 18.24 & 21.31 & 18.72 & 11.50 & 6.85 & 18.65 \\

\hdashline

\multirow{4}{*}{\bf{SentCLEME-dependent}}
& \bf{P} & 63.05 & 41.07 & 57.60 & 95.27 & 100.0 & 67.45 & 55.17 & 63.62 & 53.26 & 65.05 & 82.90 & 98.70 & 61.78 \\
& \bf{R} & 48.87 & 59.94 & 51.21 & 32.37 & 31.33 & 36.07 & 49.71 & 44.84 & 50.61 & 44.57 & 36.06 & 32.15 & 44.43 \\
& \bf{F$_{0.5}$} & 36.24 & 32.94 & 37.39 & 31.51 & 31.33 & 23.25 & 32.24 & 32.56 & 33.34 & 32.37 & 31.93 & 32.30 & 31.46 \\

\midrule

\multirow{11}{*}{\bf{CLEME-independent}}
& \bf{TP} & 318 & 488 & 392 & 17 & 0 & 33 & 272 & 196 & 339 & 204 & 69 & 24 & 214 \\
& \quad w/o LW & 388 & 596 & 487 & 22 & 0 & 39 & 338 & 248 & 420 & 255 & 85 & 32 & 262 \\
& \bf{FP} & 864 & 1382 & 1016 & 72 & 0 & 529 & 965 & 773 & 1236 & 781 & 292 & 19 & 843 \\
& \quad w/o LW & 809 & 1295 & 948 & 67 & 0 & 488 & 897 & 707 & 1136 & 781 & 272 & 18 & 787 \\                                         
& \bf{FN} & 928 & 701 & 884 & 1362 & 1393 & 1246 & 937 & 1022 & 883 & 1025 & 1266 & 1372 & 1026 \\
& \quad w/o LW & 1029 & 778 & 984 & 1497 & 1530 & 1382 & 1045 & 1129 & 990 & 1135 & 1398 & 1506 & 1136 \\
& \bf{TN} & 6629 & 6701 & 6597 & 6567 & 6560 & 6664 & 6617 & 6611 & 6793 & 6628 & 6583 & 6546 & 6680 \\

& \bf{P} & 26.90 & 26.11 & 27.85 & 19.29 & 100.0 & 5.85 & 22.00 & 20.23 & 21.50 & 20.69 & 19.02 & 56.40 & 20.22 \\
& \bf{R} & 25.53 & 41.06 & 30.71 & 1.25 & 0.00 & 2.57 & 22.52 & 16.10 & 27.72 & 16.59 & 5.14 & 1.75 & 17.23 \\
& \bf{F$_{0.5}$} & 26.61 & 28.16 & 28.38 & 4.97 & 0.00 & 4.66 & 22.10 & 19.24 & 22.51 & 19.71 & 12.35 & 7.77 & 19.54 \\

\hdashline

\multirow{4}{*}{\bf{SentCLEME-independent}}
& \bf{P} & 65.36 & 45.39 & 60.92 & 95.27 & 100.0 & 67.51 & 57.44 & 65.03 & 56.26 & 66.65 & 83.17 & 98.70 & 63.36 \\
& \bf{R} & 57.20 & 70.15 & 60.76 & 36.87 & 35.29 & 40.56 & 58.08 & 52.20 & 59.83 & 52.14 & 41.31 & 36.59 & 51.94 \\
& \bf{F$_{0.5}$} & 42.00 & 39.63 & 43.76 & 35.49 & 35.29 & 25.90 & 38.13 & 37.43 & 39.38 & 37.04 & 36.25 & 36.48 & 36.59 \\

\bottomrule
\end{tabular}}
\caption{\label{tab:results-CoNLL}
Detailed evaluation results across 13 GEC systems on CoNLL-2014.
}
\end{table*}
\begin{table*}[thbp!]
\centering
\scalebox{0.80}{
\begin{tabular}{llllllll}
\toprule
\bf{} & \multicolumn{1}{c}{\bf{Chunk 1}} & \multicolumn{1}{c}{\textcolor{red}{\bf{Chunk 2}}} & \multicolumn{1}{c}{\bf{Chunk 3}} & \multicolumn{1}{c}{\textcolor{red}{\bf{Chunk 4}}} & \multicolumn{1}{c}{\bf{Chunk 5}} & \multicolumn{1}{c}{\textcolor{red}{\bf{Chunk 6}}} & \multicolumn{1}{c}{\bf{Chunk 7}} \\
\midrule

Source & If & & not & their family then & who else & that are & willing to do that ? \\

Ref. 1 & If &  & not & their family then & who else & will be & willing to do that ? \\

Ref. 2 & If &  & not & their family then & who else & would be & willing to do that ? \\

Ref. 3 & If &  & not & from your family then & who else & is & willing to do that ? \\

Ref. 4 & If &  & not & their family , then & who else & will be & willing to do that ? \\

Ref. 5 & If &  & not & their family then & who else & will be & willing to do that ? \\

Ref. 6 & If &  & not & their family & who else & would be & willing to do that ? \\

Ref. 7 & If &  & not & their family then & who else & will be & willing to do that ? \\

Ref. 8 & If &  & not & their family , & who else & is & willing to do that ? \\

Ref. 9 & If & family do & not & help then & who else & would be & willing to do that ? \\

Ref. 10 & If &  & not & their family , then & who else & is & willing to do that ? \\

\bottomrule
\end{tabular}}
\caption{\label{tab:case-correction-independence}
A case of correction independence. We apply chunk partition to the source and all the references.}
\end{table*}

\begin{table*}[thbp!]
\centering
\scalebox{0.75}{

\begin{tabular}{lcccc}
\toprule

& \textcolor{red}{\bf{Chunk 1}}
& \bf{Chunk 2}
& \textcolor{red}{\bf{Chunk 3}}
& \bf{Chunk 4}  \\

\hline

Source  &  For not  &  use  &  car  &  .  \\
Ref. 1  &  Not for  &  use  &  with a car  &  .  \\
Ref. 2  &  Do not  &  use  &  in the car  &  .  \\
Ref. 3  &  Car not for  &  use  &    &  .  \\
Ref. 4  &  Can not  &  use  &  the car  &  .  \\

\bottomrule
\end{tabular}}
\end{table*}

\begin{table*}[thbp!]
\centering
\scalebox{0.74}{
\begin{tabular}{lcccccc}
\toprule

& \textcolor{red}{\bf{Chunk 1}}
& \bf{Chunk 2}
& \textcolor{red}{\bf{Chunk 3}}
& \bf{Chunk 4}
& \textcolor{red}{\bf{Chunk 5}}
& \bf{Chunk 6}  \\

\hline

Source  &  One person if do n't  &  have good health  &  that means  &  so many things  & 
 they could lost  &  .  \\

Ref. 1  &  If a person does n't  &  have good health  &  ,  &  so many things  &  could be lost  &  .  \\

Ref. 2  &  If one person does not  &  have good health  &  , that means they could lose  &  so many things  &    &  .  \\

Ref. 3  &  If one person does n't  &  have good health  &  , that means they could lose  &  so many things  &    &  .  \\

Ref. 4  &  If one person does n't  &  have good health  &  , that means  &  so many things  &  they could lost  &  .  \\

\bottomrule
\end{tabular}}
\end{table*}

\begin{CJK*}{UTF8}{gbsn}
\begin{table*}[thbp!]
\centering
\scalebox{0.75}{
\begin{tabular}{lcccccc}
\toprule

& \textcolor{red}{\bf{Chunk 1}}
& \bf{Chunk 2}
& \textcolor{red}{\bf{Chunk 3}}
& \bf{Chunk 4}
& \textcolor{red}{\bf{Chunk 5}}
& \bf{Chunk 6} \\

\hline

Source  &  今天  &  听天气预报说  &  今  &  天  &  还有天气  &  冷。   \\

Ref. 1  &  &  听天气预报说  &  今  &  天  &  天气  &  冷。  \\

Ref. 2  &  今天  &  听天气预报说  &  &  天  &  气还会变  &  冷。  \\

Ref. 3  &  &  听天气预报说  &  今  &  天  &  天气还会变  &  冷。  \\

\bottomrule
\end{tabular}}
\end{table*}

\begin{table*}[thbp!]
\centering
\scalebox{0.72}{
\begin{tabular}{lcccccccc}
\toprule

& \textcolor{red}{\bf{Chunk 1}}
& \bf{Chunk 2}
& \textcolor{red}{\bf{Chunk 3}}
& \bf{Chunk 4}
& \textcolor{red}{\bf{Chunk 5}}
& \bf{Chunk 6} 
& \textcolor{red}{\bf{Chunk 7}}
& \bf{Chunk 8}  \\

\hline

Source  &  所以  &  我从小到现在在这些快餐  &  &  吃饭的机会很少  &  。  &  对我来说每  & 
 次  &  饭都很重要。  \\

Ref. 1  &  所以  &  我从小到现在在这些快餐  &  店  &  吃饭的机会很少  &  。  &  对我来说每  &  顿  &  饭都很重要。 \\

Ref. 2  &  所以  &  我从小到现在在这些快餐  &  店  &  吃饭的机会很少  &  。  &  对我来说每  &  次吃  &  饭都很重要。  \\

Ref. 3  &  &  我从小到现在在这些快餐  &  店  &  吃饭的机会很少  &  ，所以  &  对我来说每 
 &  次吃  &  饭都很重要。  \\

Ref. 4  &  &  我从小到现在在这些快餐  &  店  &  吃饭的机会很少  &  ，所以  &  对我来说每 
 &  顿  &  饭都很重要。  \\

\bottomrule
\end{tabular}}
\caption{
More cases of chunk partition.
These tables are automatically generated by CLEME.
The first two cases are from JELEG-\textit{dev}, and the next two cases are from MuCGEC-\textit{dev}.
}
\label{tab:case-more}
\end{table*}

\end{CJK*}

\section{Detailed Analysis}

Table~\ref{tab:results-CoNLL} reports the detailed evaluation results of 13 systems on CoNLL-2014.
The reason behind the lower TP and FP counts of CLEME as compared to ERRANT is attributed to the chunk partition process, where overlapping edits are merged into chunks.
It is worth noting that the FN counts of CLEME are significantly lower than those of ERRANT because of their distinct definition.
While ERRANT considers FNs as the edits of references that are not identical to hypotheses, CLEME defines them as the corrected/dummy chunks of references that do not exactly match the chunks of hypotheses.
We believe that the definition of ERRANT could be problematic, as it has a tendency to overestimate the FN counts of GEC systems.
This may result in an underestimated Recall rate in turn.

\paragraph{\bf{-dependent v.s. -independent}.}
Comparing the Precision and Recall of (Sent)CLEME-independent to those of (Sent)CLEME-dependent, it is observed that the former has a slightly higher value.
This is because (Sent)CLEME-independent has the potential to overestimate the performance of GEC systems, whereas (Sent)CLEME-dependent could result in underestimating the same.
It is noteworthy that both metrics provide an upper bound and lower bound for GEC performance, respectively.

\paragraph{\bf{Corpus-level v.s. Sentence-level}.}
The precision, recall, and F$_{0.5}$ scores of sentence-level metrics are considerably higher than those of corpus-level variants.
There might be several factors contributing to this difference, but one possible explanation is that precision and recall values get affected by a limited number of challenging samples that contain numerous corrected/dummy chunks.

\section{Correction Independence}\label{app:correction-independence}

We introduce the term \textit{correction independence} to describe a pair of chunks where the correction of each chunk is not related to the correction of the other, as illustrated in Table~\ref{tab:case-correction-independence}.
Specifically, chunk 2 and chunk 4 are considered correction-dependent because the correction of chunk 2 \textit{family do} from Ref.9 must be matched with the correction of chunk 4 \textit{help then} from Ref.9.
However, chunk 6 is correction-independent with chunk 2 (or 4) since the correction of chunk 6 has no impact on the correction of chunk 2 (or 4).

\section{More cases}\label{app:more-cases}

We list more cases in Table~\ref{tab:case-more}, which involve JFLEG~\cite{napoles-etal-2017-jfleg} for English, and MuCGEC~\cite{zhang-etal-2022-mucgec} for Chinese.




\end{document}